\documentclass[journal,twoside]{IEEEtran}

\usepackage{algpseudocode}
\usepackage{algorithm}
\usepackage{optidef}
\usepackage{graphics} 
\usepackage{epsfig} 
\usepackage{mathptmx} 
\usepackage{times} 
\usepackage{amsmath} 
\usepackage{amssymb}  
\usepackage{t1enc}
\usepackage[hidelinks]{hyperref}
\usepackage{tikz}
\usepackage[caption=false,font=footnotesize]{subfig}
\usetikzlibrary{decorations.pathmorphing}
\usetikzlibrary{shapes,arrows,shadows,patterns}
\usepackage[printonlyused]{acronym}
\usepackage{nomencl}
\usepackage{mathrsfs}
\newcommand{\norm}[1]{\left\lVert#1\right\rVert}

\ifCLASSINFOpdf
  
\else
\fi
\begin{document}
%
\title{\LARGE \bf
Online Decentralized Receding Horizon Trajectory Optimization for Multi-Robot systems 
}

%
%
%

\author{Govind Aadithya R$^*$, Shravan Krishnan$^*$, Vijay Arvindh  and Sivanathan K  
\thanks{$^*$Both authors contributed to this work equally}
\thanks{This work was supported by SRM Institute of Science and Technology}
\thanks{Authors are with Autonomous Systems Lab, Department of Mechatronics, SRM Institute of Science and Technology, India, Email:\textit{(govinda\_adithya,shravan\_krishnan,vijay\_arvindh)@srmuniv.edu.in, sivanathan.k@ktr.srmuniv.ac.in}, }
\thanks{Corresponding Author: Sivanathan K}
}

%
%

\markboth{IEEE Transactions on Robotics, Under Review}
{Aadithya and Krishnan \MakeLowercase{\textit{et al.}}: Online Decentralized Receding Horizon Trajectory Optimization for Multi-Robot systems }
%



\maketitle

\begin{abstract}
A novel decentralised trajectory generation algorithm for Multi Agent systems is presented. Multi-robot systems  have the capacity to transform lives in a variety of fields. But, trajectory generation for multi-robot systems is still in its nascent stage and limited to heavily controlled environments. To overcome that, an online trajectory optimization algorithm that generates collision-free trajectories for robots, when given initial state and desired end pose, is proposed. It utilizes a simple method for obstacle detection, local shape based maps for obstacles and communication of robots' current states. Using the local maps, safe regions are formulated. Based upon the communicated data, trajectories are predicted for other robots and incorporated for collision-avoidance by resizing the regions of free space that the robot can be in without colliding. A trajectory is then optimized constraining the robot to remain within the safe region with the trajectories represented by piecewise polynomials parameterized by time. The algorithm is implemented using a receding horizon principle. The proposed algorithm is extensively tested in simulations on Gazebo using ROS with fourth order differentially flat aerial robots and non-holonomic second order wheeled robots in structured and unstructured environments.  
\end{abstract}


%
\IEEEpeerreviewmaketitle

\section{Introduction}
%
%
%
%
\IEEEPARstart{R}{ecent} advancements in trajectory planning for mobile robots have resulted in multi-robot systems as an emerging field wherein a lot of possible applications can be garnered. The collaborative navigation of multiple autonomous mobile robots is a necessity in many areas and has already been utilized for applications like collaborative transportation\cite{mora2017multi}, intersections \cite{intersection}, entertainment \cite{scholleing2012}. Multi-robots systems can be classified as a group of individual entities working together so as to maximize their own performance while accounting for some higher goals. The trajectories generated in such scenarios will have to ensure that the robots do not collide with one another and also dynamic limits of the agents are not violated. The trajectory generation process in multi agent systems has long since been done in a centralized manner wherein the trajectories are generated before hand and transmitted across to individual robots. This is a feasible approach if the environment is known and the number of robots are also known beforehand and limited in number as scalability is a huge problem in centralized methods. Recently, this has branched out to decentralized approaches that attempted to plan trajectories in known environments using a variety of different approaches. These factors may be known in case of entertainment or industrial environments, but other environments such as intersections and indoor areas are complex and dynamic and therefore   difficult to compute all details completely beforehand. Hence, it is important to utilize methods that are able to accommodate such scenarios by continuously re-planning the trajectories, accounting for the currently known factors of the environment with the re-planning accounting for unknown factors over time.  But it is important to be able to navigate through unknown environments while collaborating with other unknown number of robots. Attempts have been made to map environments and plan trajectories in such scenarios using multiple robots \cite{Regev2016icra}. Moreover, many multi-robot approaches constrain the robot to be at rest at the beginning and at the end \cite{turpin2014concurrent}, \cite{tang2018hold},\cite{Zhou2017real}, or assume that a preferred velocity \cite{mora2018cooperative} is given for the robot. Neither of these assumptions are tractable in dynamic environments wherein the end goal may be time dependent or rapidly changing. Therefore, it is important to consider the end pose at a desired time stamp but ensure that collisions with obstacles and other robots are also avoided. Besides, an essential consideration is that robots may have to pass through multiple way-points. Thus, in an attempt to progress towards it, an algorithm for online trajectory re-planning in multi-robot scenarios is proposed that takes into account unknown obstacles and robots in the environment, passing through multiple way-points and solves the trajectory generation problem in a receding horizon fashion.


\begin{figure*}
\subfloat[][]{\includegraphics[width=0.37\textwidth]{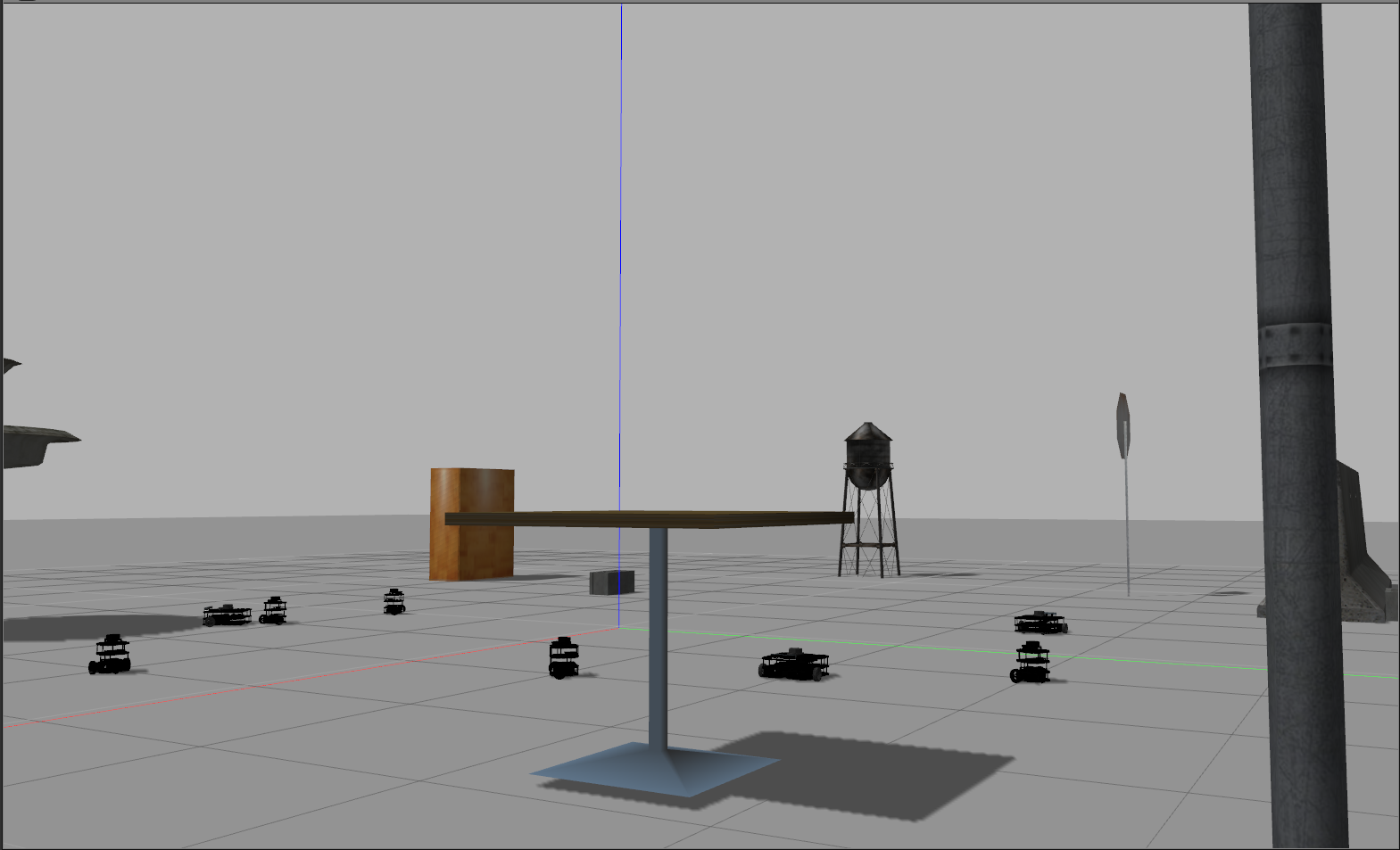}
\label{unstructured}} 
\hspace{3mm}
\subfloat[][]{\includegraphics[width=0.25\textwidth]{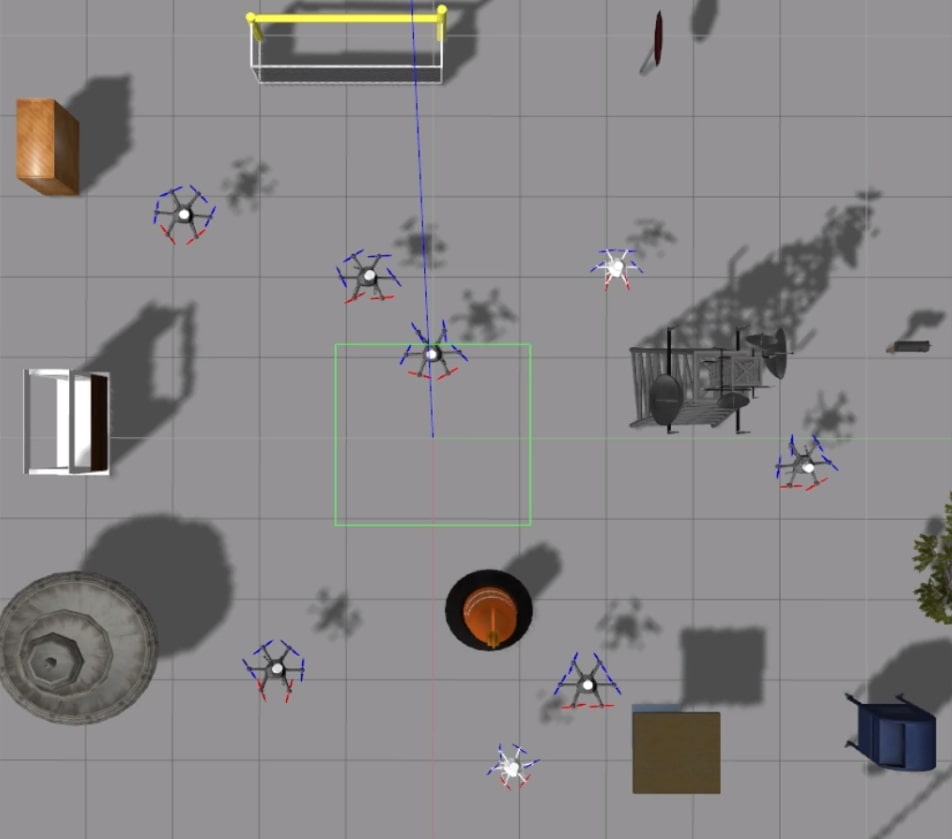}\label{aeri_unstruct}}
\hspace{3mm}
\subfloat[][]{\includegraphics[width=0.32\textwidth]{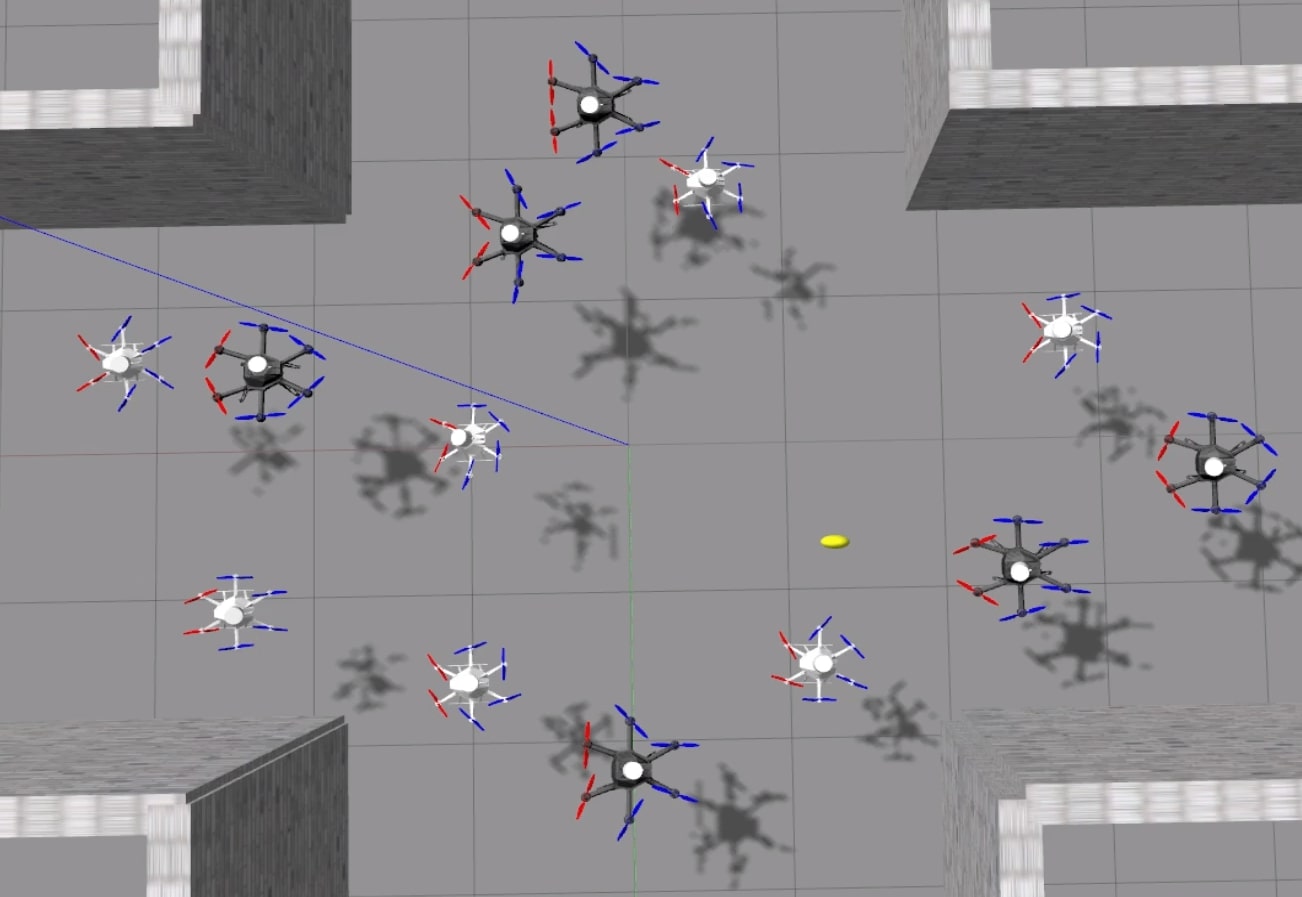}\label{13_inter}}
\caption{Snapshots while the robots are navigating across different unknown environments using the algorithm proposed in this work.}
\label{image_robots}
\end{figure*}

Our attempt is to solve this problem by using a two step process that generates collision-free convex regions that the robot can be constrained to stay within by forward-predicting other robots' positions. Using the collision-free regions, we generate smooth piecewise polynomials parametrized by time. As majority of robotic systems ranging from non-holonomic ground robots to aerial robots have been proven to be differentially flat, this time-based polynomial parametrization is valid and such methods using polynomial splines are utilized in \cite{mellinger2011minimum}, \cite{optimalwalambe2016}, \cite{Tang2009}. Obstacles in the environment are observed and stored using primitive models rather than point clouds or voxels or occupancy grids. Furthermore, a local map is utilized for querying obstacle distances using which convex regions are generated. The current data buffer from the exteroceptive sensor that showcases obstacles is augmented as soft constraint for the objective with the obstacles approximated as circles of appropriate radii.

The approach can also be interpreted as a decentralized model predictive control based trajectory generation for n\textsuperscript{th} order systems that constrains the robots to stay within a safe region at specified discrete time points in two-dimensional space.

\subsection{Contributions}
The work presents an algorithm that allows for robots to traverse across unknown environments while ensuring that the robots do not collide with other robots and/or arbitrarily shaped obstacles at arbitrary positions. In a previous work \cite{vijayroma}, we had tackled a similar problem as the one presented in this work. But in contrast to it, here the trajectories are parameterized by piecewise polynomials, a greater number of models for obstacle representation are used and much better method for utilization of the free space is proposed. Besides, obstacles are also considered for the free-space generation. Furthermore, in the previous work, we tested the algorithm for up to eight aerial robots(A fourth order differentially flat system), whereas in this work, we have tested the algorithm for a much higher number of aerial robots. Apart from that, in this work we also consider the scenario of desired end-time being unavailable and navigation through multiple waypoints.  The contributions can be stated as
\begin{enumerate}
\item A method to formulate collision-free convex regions for robots.
\item A real time decentralized trajectory optimization algorithm for multi-robot systems.
\item Method to generate maps based on primitive shapes in 2D environments based on LiDAR data under assumption of no uncertainty
\item Extensive simulation of the proposed algorithm 
\end{enumerate}

The algorithm requires minimal collaboration amongst the agents and assumes that the robots are equipped with LiDAR or depth perception sensors like RADAR. The proposed algorithm allows for the robots' end positions to be changed during runtime. A significant advantage of the proposed algorithm is continuous-time parametrization of the generated trajectory with the collision avoidance approximated at discrete intervals for other robots. Moreover, the algorithm allows for solving the problem as a convex optimization problem.

\subsection{Outline of the paper}
The related works are presented in Section \ref{Related works}. A formal problem definition and assumptions are provided in Section \ref{Problem formulation}. An overview of the method is presented in Section \ref{overview}. Section \ref{local map}  details the local map and moving volume construction and is followed by  safe region contraction while accommodating for the other robots in Section \ref{Convex}. The trajectory optimisation formulation is detailed in Section \ref{trajectory generation}. The results are discussed in Section \ref{Results} with the paper concluded in Section \ref{Conclusions}.

\section{Related Works}
\label{Related works}
Collision-free trajectory planning for multi-robots is a field that has been researched upon by many different researchers. We first discuss current methods for multi-robot trajectory generation and the review related works that provide collision-free regions for trajectory decomposition

\subsection{Trajectory generation for multi-robot Systems}
Trajectory generation for multi-robot systems can be broadly classified into decentralized and centralized approaches. Most algorithms utilize a centralized approach for multi-robot trajectory generation as centralized approaches allow for a detailed control over the complete system and complete collaboration of the underlying robots. But, recently, there has been a spur of algorithms that utilize decentralised approaches for trajectory planning as they allow for a higher scalability and real life transferability.

In \cite{turpin2014concurrent} A method for concurrently assigning robot's goal positions and trajectories were addressed for inter changeable robots(homogeneous robots). They formulated the trajectory as polynomials and generated  collision-free trajectories in obstacle-free environments. The assignment of positions was solved as a linear problem following which calculus of variations was used to  to generate collision-free trajectories. They also proposed a decentralized method wherein collision-free trajectories are planned by having the robots' transmitting their current and final positions. A constrained centralized optimization based method was proposed for formations of multiple robots using Sequential Quadratic Program by computing free regions in space-time for the robots in \cite{mora2017multi}. They plan trajectories for the overall robots formations in a forward time horizon and then utilize that trajectories and find local trajectories for the robots individually. This method requires the solution of a non convex optimisation problem which is a difficult assumption for real time implementation.
In \cite{tang2018hold}, a centralized multi-robot trajectory planner in obstacle-free environments was proposed utilizing tools from non linear optimization and calculus of variation.  Furthermore, the approach utilized a two step process for planning trajectories wherein a piecewise linear trajectory is generated based upon geometric constraints in the first step and in the subsequent step a higher order polynomial parametrized trajectory is generated. Trajectory planning is based upon a quadratic programming approach with the collision avoidance constraints enforced using separating hyperplanes. The usage of separating hyperplanes with the previously planned linear trajectory results in a very small amount of free space possible which is a conservative approximation. In \cite{robinson2018efficient}, a sequential centralized algorithm for planning trajectory in non convex environments based upon level set and orthogonal collocation was proposed. The algorithm showed good scalability but did not re-plan trajectories or have a method to account for unknown obstacles.

A centralized mixed integer programming based approach to multi-robot path planning was proposed by \cite{schouweenars2001mixed} wherein static and moving obstacle avoidance were accounted for using binary integer constraints. The method also accounted for fuel efficiency of the robots. A topological method for centralized multirobot collision avoidance utilizing braids and therein designate pairwise interactions amongst robots was proposed in \cite{mercado2017mixing}.  

Sampling based methods have also been proposed for multi-robot navigation that use discrete graph based planning for navigation \cite{Solovey2016finding}. A real time re-planning approach for multi-robot systems was proposed by \cite{fan} who utilized an A* algorithm that accounted for environmental changes and re-planned paths for the robots according to environment. The proposed algorithm is albeit computationally demanding due to A* algorithm being very expensive and A* is also a discrete graph method.  

Decentralized approaches result in trajectories for the robots being distributed incorporating constraints like lack of complete information of the systems. A model predictive control method for multi-robot navigation in polygonal based environments using non gradient vector fields and kinematic model was presented in \cite{Hedge2016}. This work was extended
in \cite{sutorius2017decentralized} where a decentralized planning for multi agent systems collaboration was proposed using polygonal based representations for convex and non convex obstacles. They utilized a hybrid method to detect collisions and utilized a switching systems to achieve the same. The method only took an immediate action for the robot to move. In \cite{van2016online}, an online distributed system is presented for multi-robot holonomic robots using alternating direction method of multipliers. The robot is formulated by a kinematic model with trajectories parametrized by B-Splines. Then a slack of the trajectory of other robots is available to generate feasible collision-free trajectories. The robots transmit amongst themselves the iterated solution from the previous iteration. Generating regions of free space for the robots was attempted using voronoi cells in \cite{2017fast}. They deflated the free space by the robot's radius and then formulated trajectories assuming the robots are initially in a collision-free areas. They utilized a receding horizon control based approach, which was formulated as a convex quadratic program.
\begin{figure*}[ht!]
\includegraphics[width=\textwidth]{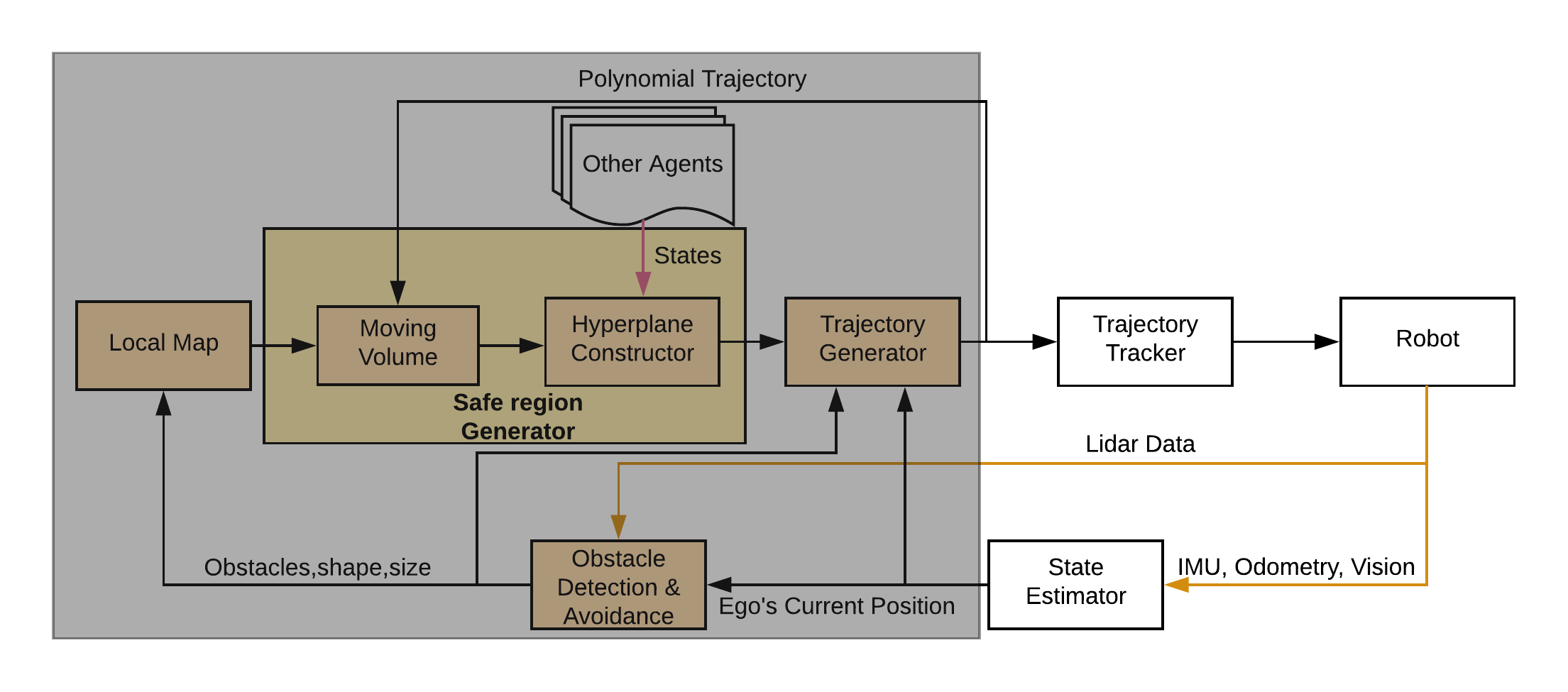}
\caption{The system overview of an individual agent. The orange arrows represent raw data,red represent data from external sources and black represent processed data and flow side. The focus of this work is the shaded portion}
\label{Overall Data flow }
\end{figure*}

Distributed collision avoidance for multi-robot systems have also been attempted \cite{Berg2008} in a method called reciprocal velocity obstacle, exploiting the concept of Velocity obstacles proposed by \cite{Fiorini1998}. This approach assumed other agents continued moving in a straight line with future collisions accounted for by relative velocities. Based on Reciprocal velocity obstacles(RVO), many velocity obstacle based approaches were proposed $\epsilon$CCA \cite{mora2018cooperative}, ORCA \cite{Jur2011} , \cite{rufli2013reciprocal}, \cite{snape2010smooth} that were intent on finding optimal velocity trajectories for robots in a distributed manner. These works attempted to overcome the drawbacks of RVO such as reciprocal dances, which was overcome using Optimal Reciprocal Collision Avoidance \cite{Jur2011}, that used linear programming. The extension of RVO for n\textsuperscript{th} order integrator dynamics was proposed by \cite{rufli2013reciprocal}. The aforementioned methods were restricted to homogeneous teams of robots and extended for heterogeneous teams of robots in \cite{bareiss2017general}. Builiding on all these works, $\epsilon$CCA\cite{mora2018cooperative} proposed a collaborative collision avoidance algorithm for non-holonomic robots with grid based environments. The planned velocity trajectories over a time horizon and re-planned trajectories whilst respecting the vehicular constraints and also accounting for potential tracking errors of the robot that is planning. The proposed method requires a grid based map and a motion plan for the robots but can be implemented in a centralized or distributed methods but linearizes the non convex set of reciprocal velocities around the current velocity.   
  
A fully distributed algorithm for navigation in unknown environments was proposed in \cite{Zhou2017real}, using positions amongst robots communicated upon requests. Furthermore, a kinematic method of prediction and incremental sequential convex optimisation for trajectory generation in a model predictive control setting was utilized. The system only considered second order models and required specified end time limits and zero derivative at the endpoints. They proposed the usage of obstacles as shapes but didn't provide a method for finding them out. Another approach for decentralized planning for aerial robots was proposed in \cite{cole2018reactive}. They used a similar method as the one presented here for isolating the obstacles but planned a trajectory that linked a high level planner and low level controller but in contrast to this work, they accounted for wind disturbances, uncertainty in the obstacles and dynamic obstacles. But one major drawback of the approach was that obstacles and robots were modeled as circles and the robots were expected to communicate amongst themselves their full trajectories. Furthermore, the algorithm only accounted for the immediate close obstacles and used second order models. A model predictive control based approach for non-holonomic vehicles in unknown environments was proposed in \cite{hoy2012collision}. The work in \cite{ferrera2017decentralized} assumed a unicycle model for the robots in a similar 2D unknown obstacle filled environment with obstacles modeled as circles. In \cite{xu2013}, a decentralized method based on artificial moment for planning in unknown environments was proposed.

Distributed re-planning for multiple robots with every robot having different planning cycles in known obstacle-filled environments was attempted in \cite{Bekris2017Safe}.The robots transmitted previously generated trajectories and planned trajectories while avoiding collisions with these trajectories and also incorporated conservative approximations to account for deviations from the transmitted trajectories. The planning of trajectories was done in a sampling based method.

It would be remiss to not look at trajectory re-planning for single robot as decentralized systems can fundamentally be looked at as an individual robot planning trajectories in dynamic environments. The recent past has seen many different methods for trajectory re-planning in unknown environments using  continuous-time representations and unconstrained trajectory optimisation. The aforementioned work was attempted for aerial robots in \cite{safe2018},\cite{2017real}. These approaches represented the trajectories by time parameterized polynomials wherein \cite{safe2018} used polynomial based representation and \cite{2017real} used B-Splines. 

\subsection{Partitioning of Space}
Decomposition of the area into available and occupied area simplifies the problem of finding trajectories. A semi definite program based approach was formulated in \cite{convexregions} for generating convex spaces in higher dimensions. They utilize a non trivial procedure for formulating the speeds and require geometric representations of the environments. Moreover the algorithm requires solving a Quadratic program(QP) followed by a semi definite program and is hence difficult to implement in real time applications. A simpler approach for convex decomposition was presented in \cite{plann} that utilized a sampling based planner of the robot thereby formulating a piecewise linear path and then expanding it outwards while ensuring obstacles do not come into the region. The proposed method for safe region is inspired by the approach presented in \cite{plann} but allows for more viable safe regions and is expanded around a point rather than a line segment. Recently, a method for partitioning space-time for autonomous robots was attempted by \cite{partition}, wherein they formulated free spaces in 2D and time separately and then generated a union of it  by trapezoid based decomposition to generate the free space-time. A method for free space generation for aerial robots was proposed in \cite{gao2018online} utilizing fast marching method with signed euclidean distance field to generate free space with trajectories parameterized by bezier curves.

\section{Problem formulation}
\label{Problem formulation}
Consider $N \in \mathbb{N} $ robots in a 2 dimensional workspace with an unknown number of obstacles and their sizes. The position of i\textsuperscript{th} robot is represented by $ x_i \in \mathbb{R}^2 $. Each of the $N$ robots has a set of desired poses(non interchangeable goal positions/waypoints). Every robot is assumed to transmit their current time stamped state to other robots in the vicinity at frequent intervals. The robots are modeled as $n^{\text{th}}$ order integrators with the states as $\textbf{\textit{x}}_i = [x_i \hspace{2mm} \dot{x}_i \hspace{2mm} \ldots \hspace{2mm} \frac{d^{n-1}x_i}{dt^{n-1}} ]$. Where $ \dot{x}_i$ is the velocity of i\textsuperscript{th} robot along the two axes, $ \frac{d^{n-1}x_i}{dt^{n-1}}$ is the $n-1^{\text{th}}$ derivative of i\textsuperscript{th} robot. The $n^{\text{th}}$ derivative is the input to the system in each dimension. A common reference frame for all the robots is a requisite as the robots share their pose estimates. It is required that the robots go from their current positions through all their desired poses as close as possible to the specified timestamps (if any). Alternatively, for all the robots, a trajectory is to be planned that ensures the robot traverses from its current position to the desired position within the specified time while not colliding with any of the other robots and/or obstacles in the environment. 

Furthermore, we assume that each robot is equipped with a rangefinding based sensor that can give the depth information of the obstacles and that the depth sensor only perceives obstacles within a sensing region. We also assume that the robots do not know the number of robots in the environment and their desired poses and just utilize the states received by them for forward-prediction.  The centralized problem can be formulated (from an optimal control/calculus of variation perspective) as  :

\begin{argmini}|l|
 {\textbf{\textit{X}}}{\sum_{i=0}^{i=N}\int_{t_1}^{t_h}\frac{d^{n-1}x_i}{dt^{n-1}}^2 + \norm{\frac{d^{n}x_i}{dt^{n}}}^2 \hspace{1mm} dt} {}{}
\addConstraint{\textbf{\textit{X}}(t_1)}{=X_0}{}
\addConstraint{\textbf{\textit{x}}_i}{\in \chi_i,}{i=0,\ldots,N}
\addConstraint{\textbf{\textit{u}}_i}{\in \upsilon_i,}{i=0,\ldots,N}
\addConstraint{x_i(t) \cap x_j(t)}{=\emptyset i,j=0,\ldots,N\hspace{1mm} \&  i \neq j}
\addConstraint{\textbf{\textit{X}}^{des}=\textbf{\textit{X}}(t_2) }{} {}
\label{Overall Problem}
\end{argmini}

The constraints on the overall system are:
\begin{enumerate}
\item The current state of all the robots
\item The states remain within the feasible set of the respective robots
\item The positions of robots at any time from $t_1$ to $t_2$ should not coincide.(Inter-Agent Collision Avoidance)
\item The position of an obstacle and a robot should not overlap (Obstacle Avoidance)
\item The robots reach its desired end poses 
\end{enumerate}

The solution that is proposed in this work attempts to decentralize the Problem in \eqref{Overall Problem}.

\section{Method Overview}
\label{overview}

The overall system consists of a state estimation, local Map, safe region generator, trajectory optimizer. The data flow of the system is represented in Fig \ref{Overall Data flow }. Using the data from the LiDAR, exploiting the pattern of reflection, distance to obstacle, the obstacle's shape and size is inferred. The obstacle's shape, size and center are stored. Using these obstacles and specified time points, by incrementally searching the farthest distance along each direction where no obstacles are found, collision-free convex regions are formed. This is done at discrete time points based upon a discretization $\tau$  The algorithm upon receiving the robots' states,  forward-predicts the other robots' trajectories. Based upon the forward-prediction, the convex regions are modified. We restrict the robot's position at this time point to be within the formulated region and plan a trajectory. The trajectory optimization problem is then formulated to minimize a robot's n\textsuperscript{th} and n-1\textsuperscript{th} derivatives' squared integral over the specified time horizon. This optimization problem is solved by generating multiple trajectory segments with each segment having polynomials parametrized by time along each dimension. A part of this trajectory is then applied by the robots after which the trajectory is re-planned.

Thus, during each planning interval, the following steps occur on each robot :

\begin{enumerate}
\item State estimation of the robot
\item Obstacle detection and positioning
\item updating the local map
\item Safe region generation 
\item Other robots' forward-prediction
\item Safe region contraction
\item Trajectory optimisation
\end{enumerate}

Looking through the steps, it is obvious that a considerable amount of time is required to generate the trajectory. It would be efficient if the trajectory generation can be achieved quickly \begin{algorithm}
\caption{Real Time multi-robot Trajectory Optimization}
\label{Real_time}
\begin{algorithmic}
\State \textbf{given:}
\State \hspace{3.5mm} Agents' current state and size 
\State \hspace{3.5mm} Number of robots (N)
\State \hspace{3.5mm} Current pose estimate of the robot
\State \hspace{3.5mm} Timestamped desired poses of the robots

\State \textbf{Trajectory optimization:}
\State Obstacle detection and positioning as detailed in Section \ref{local map}
\State Safe region contraction as detailed in Algorithm \ref{obs} 
\State Trajectory optimization solving optimization problem detailed in Section \ref{trajectory generation}
\end{algorithmic}
\Return Trajectory, Obstacles

\begin{algorithmic}
\State \textbf{given:}
\State \hspace{3.5mm} Trajectory 
\State \hspace{3.5mm} Obstacles
\State \hspace{3.5mm} Local map
\State \textbf{Moving Volume:}
\State Obstacle insertion into local map as detailed in Section \ref{local map}
\State Moving volume along the trajectory 
\State Safe collision-free regions
\end{algorithmic}
\Return Safe collision-free regions, Local map
\end{algorithm} from the acquisition of state data. To aid in achieving that the 
following method is utilized(inspired by the real time iteration process for Non Linear Model predictive control \cite{diehl2005real}, \cite{diehl2002real}).

\begin{figure}
\includegraphics[width=0.5\textwidth]{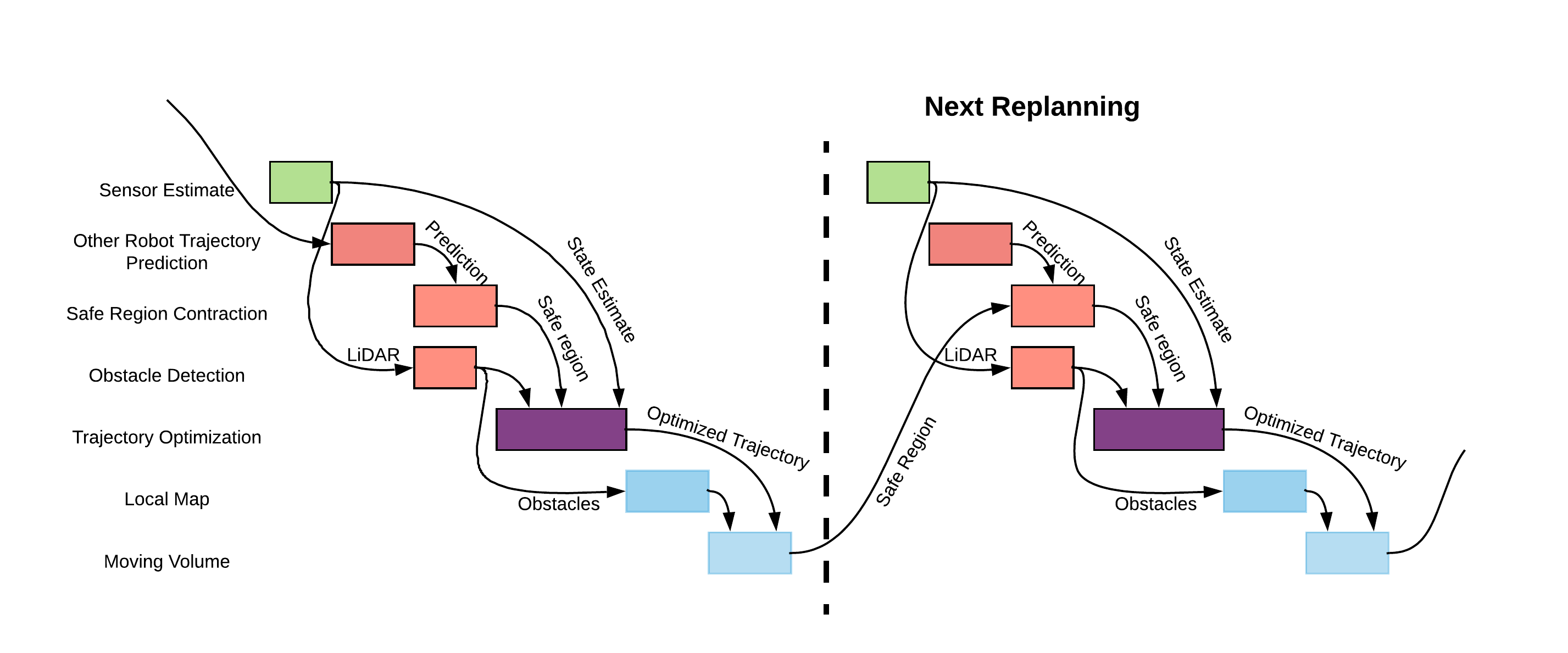}
\caption{An illustration of the steps for trajectory optimization with the subsequent data that is being transmitted. The illustration was inspired by \cite{vukov2015embedded}}
\label{RTI}
\end{figure}

Using the data from the laser range finder, obstacles' shape and sizes are found and augmented into the objective function as an obstacle cost with only the nearest obstacles to the previous trajectory added into the cost function. In the time between two subsequent odometry data, the current obstacle observation from the LiDAR is inserted into the local map if it is determined to be a new obstacle or refinement of an already known obstacle. Furthermore, assuming knowledge of trajectory re-planning rate and perfect trajectory tracking, the local map is moved appropriately along the trajectory planned in the previous iteration and at specified time points, safe regions are formed from the obstacle in a moving volume that allows for collision-free navigation. This collision region is contracted if any of the other robots come into the free region in the next re-planning of the algorithm. Otherwise the region is used directly as the safe region. This reduces the time required for one re-planning cycle as only the obstacle detection and safe region contraction have to be done and moreover, these processes are independent of each other and can be run in parallel. Thus, the algorithm uses an incrementally updated local knowledge of the environment and proceeds as shown in Algorithm \ref{Real_time}. An illustrative example of the data flow and sequence of events and interchange of data across the different re-planning is shown in Fig. \ref{RTI}.

Furthermore, the method also allows for an easy initialization of trajectory optimisation based upon just the first observation from LiDAR data. 

\section{Local Map}
\label{local map}
A  local map representation is implemented for obstacles in the environment. The robot is assumed to be employed with a 2D laser range finder, whose data is utilized to find the shape of the object and its center. The data from the LiDAR provides the distance to the obstacles.

\subsection{Obstacle detection}

The detection of reflection pattern is achieved by using a search through the data that is available in the LiDAR, that is not infinity(no reflection back). The  set of distance point reflections together  form a single obstacle as long as the reflected points are not separated by a another set of non reflecting points. As the resolution of any LiDAR is known beforehand, using the resolution and distance to the obstacle from the robot's current position, the reflected point's position can be calculated by :

\begin{equation}
\begin{split}
x_{obs}= l*cos(\theta) + P_{x_{ego}} \\
y_{obs}= l*sin(\theta) + P_{y_{ego}}
\end{split}
\label{distance}
\end{equation}
 
Thus, if a set of distance points form a straight line, the obstacle is modeled as a square with it's length determined by distance points. Similarly, if the data formulates a curve , an appropriate circle can be fit. This results in a primitive but a simple structure of the robot's obstacles to be incorporated. The selection of circle, square and rectangle is motivated by the reason that higher sided convex polygons can be approximated accurately by circles of appropriate radii. While a square can be used to accurately approximate  majority of day to day objects and is computationally simple to approximate from a partial observation. Moreover, a triangle representation is not used unless incremental scans are matched, differentiating a triangle and a rectangle is difficult from three points. To compensate for the motion of the robot between two subsequent readings, the LiDAR data is separated into segments and in each segment, the median time is calculated and used to find the position of the robot corresponding to that segment from the previously generated trajectory. This position is utilized  in Equation \ref{distance} 

\subsubsection{Recursive Obstacle shape detection}

After separating the obstacles, the first, middle and end distance to obstacles are transformed to  positions and using the positions of three reflections, a circle fitting those three points is found as a circle fitting three points is unique. The circle and it's center is found by:

\begin{equation}
\begin{split}
obs_{x_{cen}}=\frac{A_1^x + A_2^x + A_3^x}{B} \\
obs_{y_{cen}}=\frac{A_1^y + A_2^y + A_3^y}{B} \\
\end{split}
\label{circle}
\end{equation}
\begin{equation*}
\begin{split}
\text{where, } A_1^x= (_1x_{obs}^2+ _1y_{obs}^2)(_2y_{obs}-_3y_{obs}),\\ A_2^x= (_2x_{obs}^2+ _2y_{obs}^2)(_3y_{obs}-_1y_{obs}), \\ A_3^x= (_3x_{obs}^2+ _3y_{obs}^2)(_1y_{obs}-_2y_{obs}) \\
A_1^y= (_1x_{obs}^2+ _1y_{obs}^2)(_2x_{obs}-_3x_{obs}),\\ A_2^y= (_2x_{obs}^2+ _2y_{obs}^2)(_3x_{obs}-_1x_{obs}), \\ A_3^y= (_3x_{obs}^2+ _3y_{obs}^2)(_1x_{obs}-_2x_{obs}) ,
\end{split}
\end{equation*}
\begin{equation*}
\begin{split}
B= 2(_1x_{obs}(_2y_{obs}-_3y_{obs})+ _2x_{obs}(_3y_{obs}-_1y_{obs}) \\ +_3x_{obs}(_1y_{obs}-_2y_{obs}))
\end{split}
\end{equation*}

If $D$ returns a zero, or the radius of the circle is greater than a threshold value, the object is modeled as a square. Else, the middle value between the two selected points is taken and circle fitting is done again. If D returns a 0 or radius is greater than a threshold, we model the obstacle as a triangle consisting of the initial 3 points. The algorithm for shape detection is run recursively using this paradigm to check for triangles. The number of iterations, in our experiments, has been restricted to five at the maximum as the obstacle shape can be found.  

This method while being primitive, allows for simpler obstacle representations for a robot functioning in a 2D environment. Hence, the obstacles can be easily stored as with their center and sizes(radii for circular obstacles, center and corner points for squares and triangles).

\begin{algorithm}
\caption{Obstacle Detection}
\label{obs}
\begin{algorithmic}
\State \textbf{Given:}
\State \hspace{3.5mm} Obstacle distance points
i=1
\While{B!=0}
\State select the first, middle and last obstacle reading
\State solve Equation \ref{circle}
\If{B==0 || radius >= 100 }
\If{i==1}
\State Square obstacle 
\Else
\State rectangle obstacle
\EndIf
\State \textbf{break}
\EndIf
\EndWhile

\end{algorithmic}
\Return Obstacle shape
\end{algorithm}

\subsubsection{Obstacle modeling}
From the detected shape, the size of the obstacle can be accommodated for easily, 
\begin{itemize}
\item \textbf{Square/ rectangle:} Formulating a square or a rectangle is straight forward as the lengths of the squares and rectangles can be found from the lines representing their perimeter from an arbitrary line $ ax + by = c $, for a parallel line $a$ \& $b$  values are the same and only the $c$ value changes. Hence, using two point form, the equation of the line segments bounding the obstacle can be found. Moreover, from the edges of each line, perpendiculars can be constructed away from the robot and thereby constructing the box approximation. The center of the obstacle can be found from thereon easily.

\item \textbf{Circle} A circle can be easily stored  using the circle's center and radius.

\end{itemize}

\subsection{Local Map}
To keep track of the local map, we utilize a hash table based local map that allows for easier search and insertion of obstacles. The local map is robot centric around it's current position. The obstacles' center's position, shape and size are stored with the hash table indices formed by rounding off the obstacles center's distance to robot position to the closest natural number.

\begin{figure}
\subfloat[][]{\includegraphics[width=0.5\textwidth]{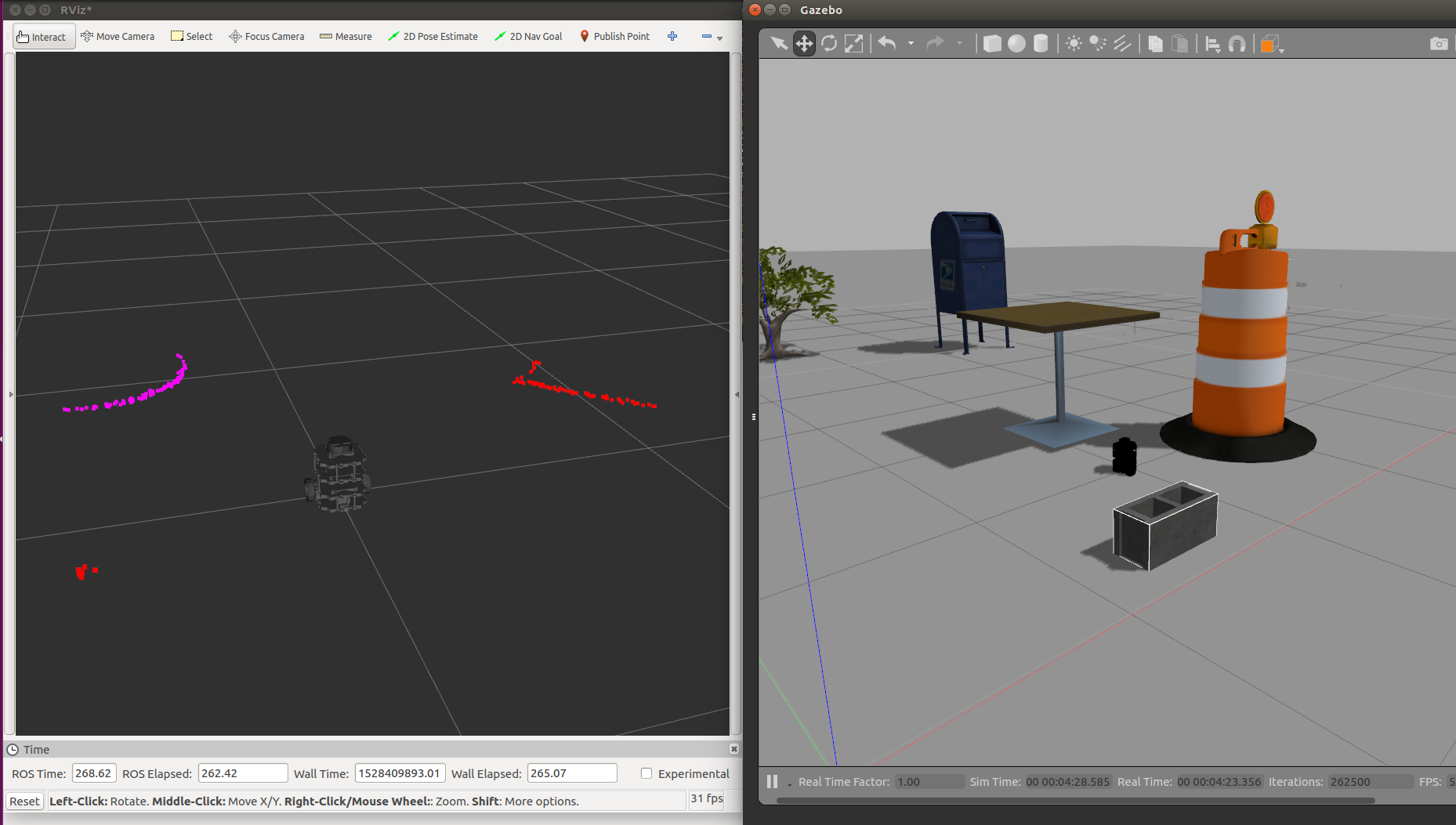}\label{LiDAR1}}
\caption{The LiDAR data visualised using Rviz. Using these LiDAR data, the obstacle shape and size are accommodated for and stored. The LiDAR is visualizing obstacles in an unstructured environment. From the image, it can be noted that the cinder block is occluded. In the subsequent observation, more of the cinder block is observed and the size of the obstacle is changed appropriately. It should be noted that the stand for the table is observed as a clutter of points and thereby modeled as very small rectangle}
\label{LiDAR}
\end{figure}

\subsection{Obstacle Insertion}
Obstacle insertion into the local map is done based on the distance to the obstacle. To restrict redundant obstacle insertion, if two obstacles have a center within a region of their lengths/radius, the two obstacle centers are appended together. As an example, in Fig \ref{LiDAR1}, the algorithm will measure the obstacle( the cinder block) as a rectangle. As the robot moves further towards the obstacle, the obstacle's subsequent measurement models it as a rectangle of different size again. Thus, during measurement insertion, two rectangles are available with different sides and they are appended together to formulate a bigger rectangle. Similar process is done for other homogeneous obstacle shapes.

The insertion for heterogeneous shape is performed by a comparison of which of the two measurements perform a better fit of the given points and that one is utilized.

\subsection{Moving Volume based Safe region}
\label{safe region}
To construct appropriate safe regions, a moving volume of obstacle map is created about the previously generated trajectory; knowing the frequency of operation. At specified time	points, along the time horizon $t_h$(based upon the discretization $\tau$), a moving volume is formulated using the local map via the distance from the position of the robot to the obstacle's center and stored.

Using the moving volume, at the discretized points, the closest obstacles to the robot in all directions is sampled incrementally. This is done to ensure that available free space is maximized while also ensuring the generated region is convex. Using these discrete obstacle points and center of obstacle, a line joining these two points and intersecting the obstacle is found and the intersection point is stored. Similarly, for all obstacles, the intersection points are found and lines are formulated joining all the obstacles. These lines are formulated by extending the line that the point of intersection is on if the obstacle is a rectangle or formulating a tangent if the robot is a circle. Using the property that the intersection of hyperplanes or halfspaces is convex \cite{boyd2004convex}, the generated area is a convex polyhedron

\section{Safe Region with other Robots}
\label{Convex}
For the formulation of the safe regions for the robots to plan trajectories within, other robot's trajectories are forward-predicted utilizing a sliding window of states that have been received by the robots so far. 
\subsection{Forward Simulation} 
The forward-prediction of the robots during the first iteration when the first transmitted state has just been received is done by using a constant acceleration approximation:

\begin{equation}
P_i(t) =  P_i(t_\delta) + v_i(t_\delta)(t-t_\delta) + a_i(t_\delta)(t-t_\delta)^2
\label{prediction}
\end{equation}

where $ t_\delta $ is the time stamp of the robot's transmitted state.

The forward simulation is done for specified time horizon $t_h$  based upon the discretization $\tau$.

In the subsequent iterations, using the transmitted states in a moving horizon formulation, we use least squares to predict time parameterized polynomials for the other robots' states in 2D. But in contrast to our trajectory optimization framework, we only use a single polynomial for the trajectory prediction. Assuming that in the time horizon $t_1$ to $t_2$, $K_{states}$ different measurements of other robots' states are available, this results in(We use boldface for the polynomial estimated states and normal face for transmitted states):
\begin{equation}
\textbf{P}_i(t)= \sum_{j=0}^{5}\alpha_j t^j
\label{decision with degree}
\end{equation}

\begin{equation}
\mathcal{D}_i = [\beta_0 \hspace{1mm} \beta_1 \hspace{1mm} \beta_2 \hspace{1mm} \cdots \beta_{5}]^T
\label{per_vehicle_prediction}
\end{equation}

The objective for the trajectory prediction per robot is:

\begin{mini}
{\mathcal{D}}{\int_{t_1}^{t_2}  \norm{\dddot{\textbf{P}_i(t)}} ^2  \hspace{1mm} dt+ \sum_{i=1}^{K_{states}}\norm{\textbf{S}_i-S_i}^2}{}{}
\label{predict_objec}
\end{mini}

where, $S$ and $\textbf{S}$ represent the states of position, velocity and acceleration. The first term is a jerk smoothness regulator to ensure the jerk does not change drastically over the time horizon. Due to the end time being known($t_h$), cost can be analytically integrated and formulated. The second term is the estimation residue error. This results in the problem to be formulated as linear least squares:

\begin{mini}
{\mathcal{D}}{\norm{A_{smooth}\mathcal{D}}^2 + \sum_{i=1}^{K_{states}}\norm{A_{est}\mathcal{D}-S_i}^2}{}{}
\label{predict_ls}
\end{mini}

This has a closed form solution \cite{boyd2004convex}. Using the polynomial at discretized time points depending on $t_{delta}$ and $t_h$. Moreover, because the robots do not have an unique id, we use the previously predicted trajectory and the current transmitted state to match the robots and their ids. 
\subsection{Accommodating the size of robots}
\label{size}
At each of the discretized time points, utilizing the transmitted size of the robots, we formulate regions depending on the transmitted data. For simplicity, if three numbers are sent, the robot is modeled as a cuboid or a cube. If two numbers a cylinder and a single number, a sphere. In the case of three numbers, robots are modeled as a square of diagonal of largest side, thereby allowing the robot to rotate freely. In case of two or one, the robot is modeled as a circle. A robot inflated to its size is represented as:
\begin{equation}
\mathsf{R}_i = 
\begin{cases}
A_i P_i(t) \leq B & Square \\
\norm{{P}_i(t)-\textbf{P}_i(t)} \leq r & Circle
\end{cases}
\end{equation}

where $B $ is formulated as $P_i(t) \pm \sqrt[]{2}\max(l)$ and $r$ is the radius of the robot

\begin{figure*}
\subfloat[][]{\includegraphics[width=0.5\textwidth]{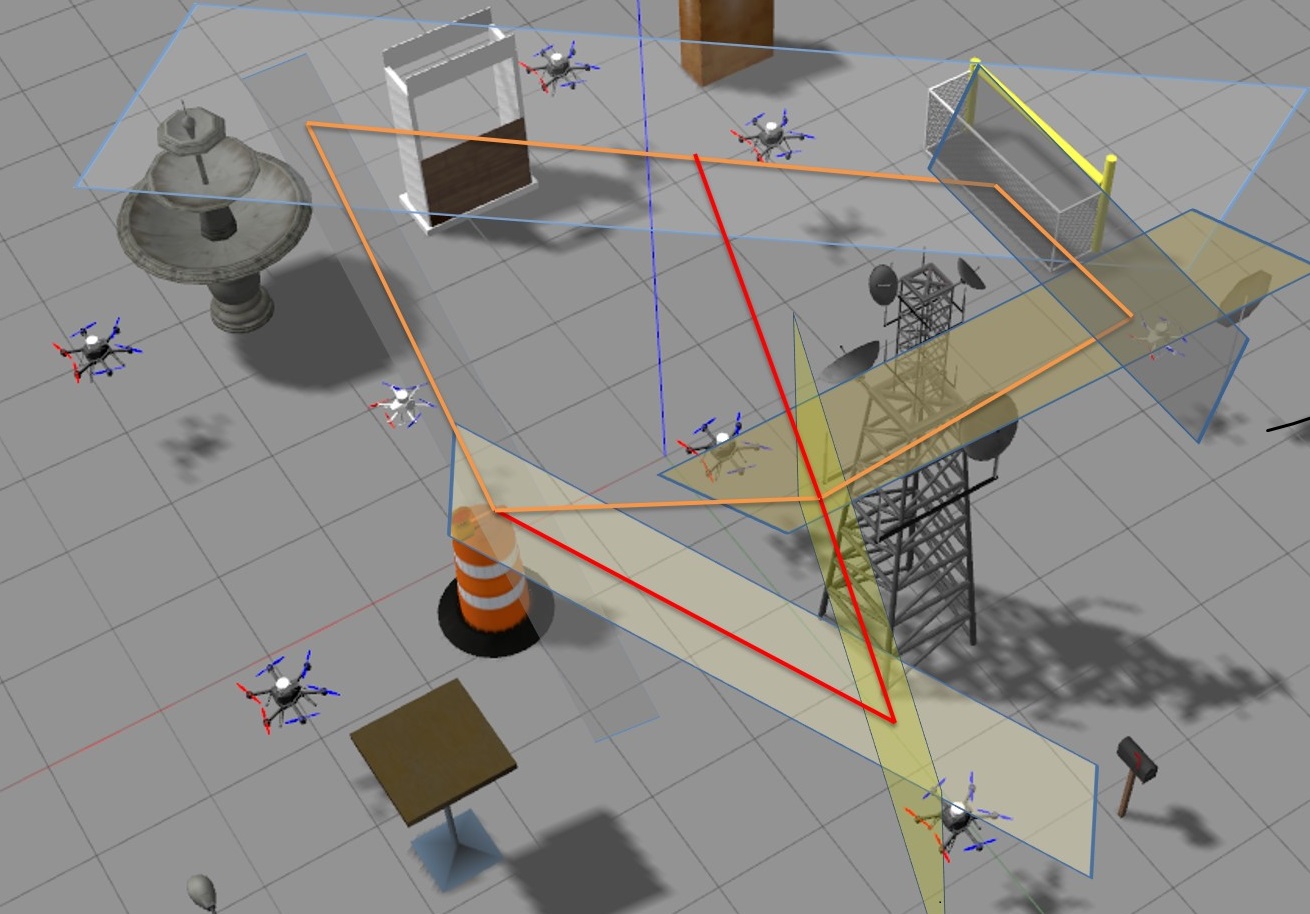}
\label{unstruct safe region}} 
\subfloat[][]{\includegraphics[width=0.5\textwidth]{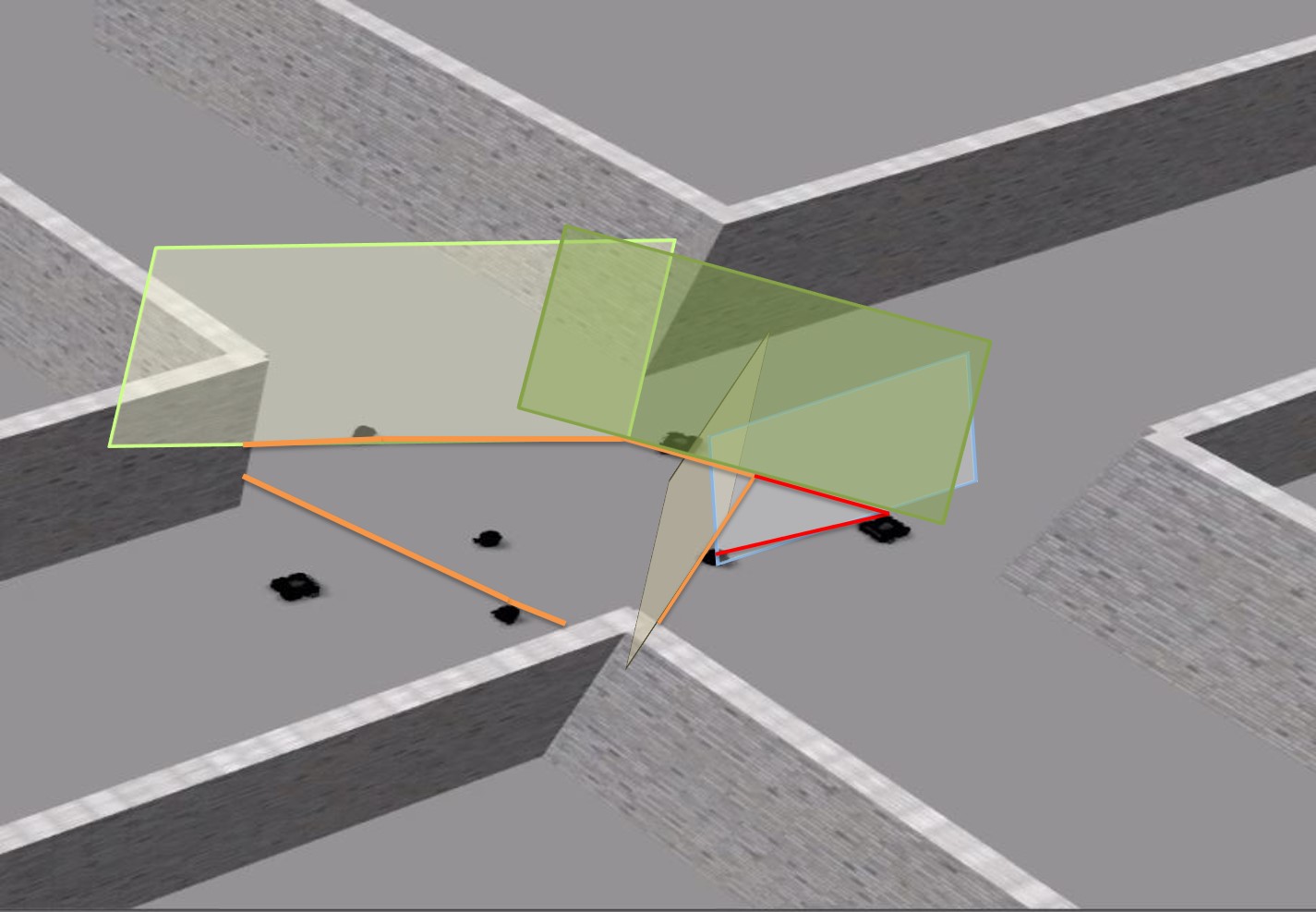}\label{Intersection safe region}}
\caption{The safe region generated by the intersection of hyperplanes for center robots. The interior of the orange shaped pentagon is the safe region  Fig \ref{unstruct safe region} shows the sage region for center AscTec Neo that at inside of the safe region and Fig \ref{Intersection safe region} shows the safe region for the Turtlebot Burger located at center of the pentagon. The red line segments in both images show the contraction of the safe region due to the presence of a robot that comes into the safe region. }
\label{convex_region}
\end{figure*}

\begin{algorithm}
\label{euclid}
\caption{Safe Convex Region}
\begin{algorithmic}
\State \textbf{Given:}
\State \hspace{3.5mm} Agents' current state and size 
\State \hspace{3.5mm} Number of robots (N)
\State \hspace{3.5mm} Moving Volume at specified time points
\For{t=$t_\delta$ to $t_h$}

\For{$i=1$ to N }
\State 
\If {data from robot received for the first time}
\State $P_i(t) =  P_i(t_\delta) + v_i(t_\delta)(t-t_\delta) + a_i(t_\delta)(t-t_\delta)^2$
\Else
\State Predict position of the robot by solving \eqref{predict_ls} per robot
\EndIf
\State Approximate size of the robots according to \ref{size}
\If {$P_i$ within Moving volume } 
\State Contract the Moving Volume
\State Get the intersection of hyperplanes
\Else 
\State Keep Moving Volume unchanged
\EndIf
\EndFor
\EndFor
\end{algorithmic}
\Return Safe Regions
\end{algorithm}

\subsection{Hyperplane}
Considering a time $t_\tau $ between $t$ and $t_h$, supporting hyperplanes are formulated for all $\mathsf{R}_i$ that are within their appropriate moving volumes and as $\mathsf{R}_i$ convex and a supporting hyperplane hence, exists. \cite{boyd2004convex}. A support hyperplane for $\mathsf{R}_i$ can be formulated as   
\begin{equation}
\eta_i(\mathsf{R}_i) \leq \eta_i(\mathfrak{r}_i^0)  
\end{equation}
where $\mathfrak{r}_i^0$ is the boundary of the set $ \mathsf{R}_i $
 
Using the method described in Section \ref{safe region} for finding an intersection point, the appropriate intersection point is formulated using the previously generated trajectory, thereby allowing to construct the hyperplane easily.

The intersection of all the support hyperplanes is convex polyhedron as an intersection of hyperplanes is convex \cite{boyd2004convex}. We constrain the robot to remain within the generated polyhedron at the discretized time points with polyhedron at each time point represented as $\mathcal{H}(t_{disc}) \leq h $. But as this region constrains the overall robot as a point we subtract the robot's dimensions from the $h$ to constrain the robot to be within the region(the convex regions at specific time points). As the robot's size is modeled invariant to rotation, contracting the safe region can be done easily. Fig \ref{convex_region} shows the generated safe region in intersection-like and unstructured environments.
  
\section{Trajectory Generation} 
\label{trajectory generation}
The generation of trajectory by the robot can be formulated as an optimization problem that tries to optimize the smoothness of the trajectory while ensuring that robots do not collide with one another.

\begin{argmini!}
  {\textbf{\textit{x}}}{C_{int} + C_{final} + C_{collision} }{}{}
  \addConstraint{\textbf{\textit{x}}(t_1)}{=x_0}{}
  \addConstraint{\sum_{t_{disc}=1}^{\frac{t_h}{\tau}}}{\mathcal{H}(t_{disc}) \leq h}{ ,\hspace{2mm}  t_1 \leq t \leq t_2 }
  \addConstraint{\underline{\ddot{x}}\leq}{x \leq \bar{\ddot{x}}}{}
  \label{cost functional}
\end{argmini!}

The above mentioned problem is continuous-time problem, therefore infinite dimensional problem. To counteract this, we formulate the trajectory in each dimension as a Uniform  B Spline parametrized by time. This time parameterized B spline results in a piecewise polynomial and can be represented as:

\begin{equation*}
x(t)= \sum_{k=1}^m B_{i,n}\alpha_i
\end{equation*}

\begin{equation*}
B_{i,n}= \frac{t-t_i}{t_{i+1}-t_i}B_{i,n-1} + \frac{t_{i+1}-t}{t_{i+1}-t_i}B_{i+1,n-1} 
\end{equation*}

\begin{equation}
\begin{split}
D = [\alpha_0 \hspace{1mm} \alpha_1 \hspace{1mm} \alpha_2 \hspace{1mm} \alpha_3 \hspace{1mm} \ldots \hspace{1mm} \alpha_m  ] \\
\end{split}
\label{per_robot}
\end{equation}

Where $m$ is the number of control points. In uniform B splines of degree $l$, $l$+1 control points are the only set of points that affect the polynomial. These $l+1$ control points can be isolated as the points that are before and after the specified knot vector in which the piece exists. Furthermore, B splines are continuous and thus, continuity can also be controlled appropriately selecting the polynomial knot points.

\subsection{Objective Function}
$C_{int}$ is the integral cost functional that specifies the objective for the derivative over the integral. $C_{final}$ is the cost at the end of the time horizon. $ C_{collision} $ is the collision cost for static obstacles along the trajectory. 

\subsubsection{Derivative Cost}
Derivative penalty utilized is square of the integral of n\textsuperscript{th} derivative and n-1\textsuperscript{th} squared over time horizon t\textsubscript{h}:
It is represented as:
\begin{equation}
C_{int}=\int_{t}^{t_h} Q_{\text{n-1\textsuperscript{th}}}\norm{\frac{d^{n-1}x}{dt^{n-1}}}^2 + Q_{\text{n\textsuperscript{th}}} \norm{\frac{d^{n}x}{dt^{n}} }^2 \hspace{1mm} dt 
\label{Derivative Cost}
\end{equation} 

where $Q_{\text{n\textsuperscript{th}}}, Q_{\text{n-1\textsuperscript{th}}}$ are tuning weights for the objective.

As the time horizon is known before hand and the initial time and position are known, this cost has a closed form solution\cite{bsplinemat} that results in it being reformulated as a Quadratic Objective with the decision vector as:
\begin{equation}
D^T H(t+t_h) D
\end{equation}
With $H(t+t_h)$ formulated by integrating Eq. \eqref{Derivative Cost} and substituting $ t$, $t_h $ and separating according to the coefficients of the polynomial.

\subsubsection{End Cost}
We add an end point quadratic cost for the final position along the trajectory as a soft constraint for two reasons. One, to allow the robot to plan appropriate trajectory if a robot or an obstacle is occupying or blocking the path directly to the end point. Two, in scenarios where the robot's end pose's time stamp is beyond it's trajectory planning horizon, this cost tries to drive the robot as close as possible to end pose, while ensuring the dynamic limits are not violated by the hard dynamic constraint

\begin{equation}
C_{final}=(x_{des}-x(t_h))^2 Q_{final}
\end{equation}

The final position alone is penalized but if required additional penalties on velocity, acceleration may also be added. This cost can be reformulated with respect to the decision variables resulting in:

\begin{equation}
D^T H(\text{Fin}) D + F(\text{Fin})^T D
\end{equation}

\subsubsection{Collision Cost}
The generated trajectory should be collision-free with respect to the obstacles that are within the safe region(newly observed obstacles). We utilize the following penalty for the avoiding such obstacles:

\begin{equation}
C_{collision}=\int_{t}^{t_h} Q_{Obs}c(x(t))v(t)\hspace{1mm} dt 
\label{Obstacle Cost}
\end{equation} 
Where
\begin{equation}
c(x) = \frac{x(t)-x_{obs}}{\exp^{K_p(d(x)-\rho)}d(x)} 
\label{Pieceswise obstacle}
\end{equation}

\begin{figure}
\label{collision_figure}
\subfloat[][]{\includegraphics[width=0.5\textwidth]{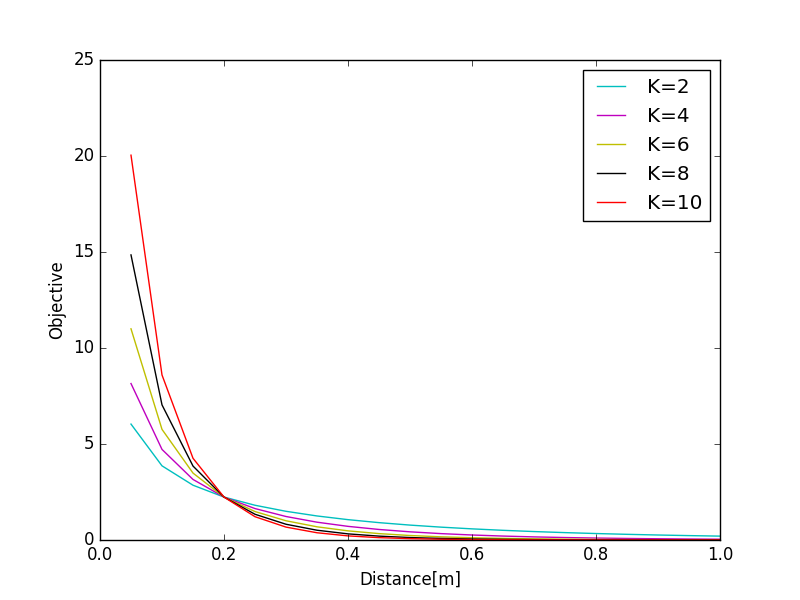}\label{negative}} 
\caption{The collision cost as shown in Equation \ref{Pseudo Obstacle Cost} for different smoothness parameters and how the cost function ensures that collision ensures collision avoidance with a threshold of 0.2}
\end{figure}

Here $d(x)$ is euclidean distance to each obstacle. Similar cost functions have been utilized for collision avoidance for autonomous cars \cite{pmpc}, aerial robots \cite{rob}. $ K_p$ is a smoothness tuning parameter that allows to increase or decrease the smoothness of the collision cost. 

The collision cost objective is the same as the one we had used in \cite{shravanicra} and can be integrated analytically and results in the following closed form solution	
\begin{equation}
\begin{split}
C_{collis}(x,obs)= Q_{Obs}[\exp^{K_p(d(x)-\rho)}\frac{-1}{K_p}]_{t}^{t_h}
\end{split}
\label{Pseudo Obstacle Cost}
\end{equation} 

To allow for faster and efficient optimization, the obstacle cost is replaced with a quadratic approximation around the previous optimized trajectory. 

\begin{equation}
\begin{split}
C_{collision}= Q_{Obs} ((x-x_{prev})^2 C_{collis}(x_prev,obs)^{''} + \\  (x-x_{prev})C_{collis}^{'}(x_prev,obs) + C_{collis}(x_{prev},obs)) 
\end{split}
\label{quadratic_approx}
\end{equation}

where,\begin{equation*}
\begin{split}
 C_{collis}^{''}(x_prev,obs) \hspace{2mm} \text{and} \hspace{2mm}  C_{collis}^{'}(x_prev,obs) 
\end{split}
\end{equation*}
are the second derivative and first derivative with respect to $x(t)$ respectively.

\begin{figure}
\includegraphics[width=0.5\textwidth]{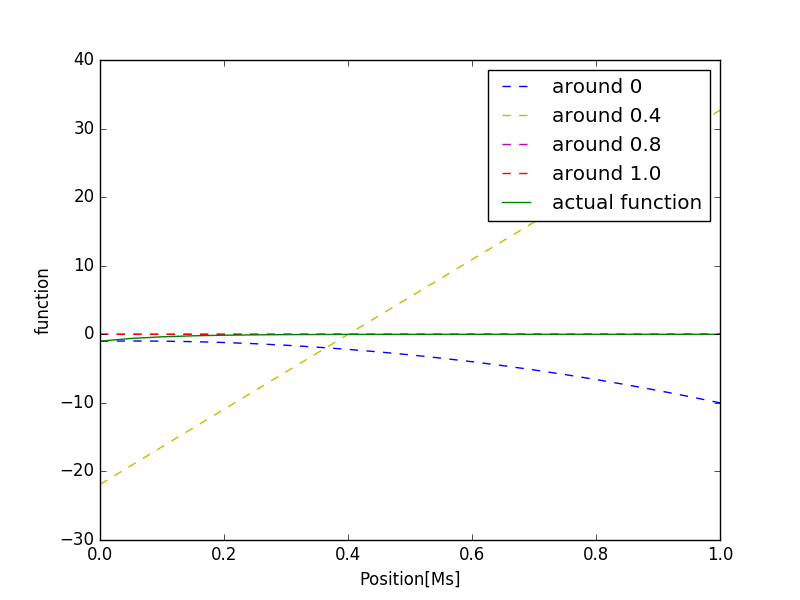}
\caption{The quadratic approximation of the closed form collision integral around different points with a time horizon of four seconds with a smoothness parameter of 10}
\label{approximation}
\end{figure}

The approximation is shown in Fig \ref{approximation}. The approximation allows to formulate the overall collision cost as a quadratic objective, resulting in:

\begin{equation}
D^T H(\text{Obs}) D + F(\text{Obs})^T D
\end{equation}

To reduce the number of obstacles for collision checking, only the obstacles that are within the safe region are augmented into the objective function for collision as the remaining is accounted by the safe region.

\subsection{Constraints}
The trajectory optimisation is constrained by the derivatives of the trajectory remaining within their feasible limits(Dynamic Constraints), the positions of the robots being within the convex region and the positions traveling through the waypoints.

\subsubsection{Waypoint constraints}
The trajectory has to also pass through the given time stamped poses along the trajectory, This results in linear equality constraints on the trajectory.

\begin{equation}
A_{way}D=P
\end{equation}
where $P$ is the stack of poses at their time.

Moreover, as the end pose's time is given in this scenario, the optimization problem, minimizes the cost only for the specific time horizon but also ensures that the robot's planned trajectory reaches the end goal at the desired time stamp. 

\subsubsection{Convex Region}
We require that the generated trajectory also remains within the feasible convex region generated at the specific time samples. This constraint is formulated as 
\begin{equation}
\sum_{t_{disc}=1}^{\frac{t_h}{\tau}}\mathcal{H}_i(t_{disc}) \mathcal{T}_i D \leq h
\end{equation}

where $\mathcal{T}$ is the map from the B spline coefficients to the positions. As both $\mathcal{H} \& \mathcal{T} $ are linear with respect to the B spline coefficients, this results in a convex constraint.

\subsubsection{Dynamic constraints}
Dynamic constraint on the robot is an infinite dimensional and hence we apply the constraints at specific points on the trajectory. This adds inequality constraints on the system

\begin{equation}
\sum_{i=1}^n \underline{d} \leq  A_{dyn}D \leq \bar{d}
\end{equation}

where $n$ is the number of discrete points wherein the constraints are added and $\underline{d} \& \bar{d}$ represent the minimal and maximal limits of the derivatives.

The resulting optimization problem can be formulated as a Non Linear Program

\begin{mini}|l|
  {D}{D^T H_{net} D + F_{net}^T D   }{}{}
  \addConstraint{A_{Cont}D}{=0}{}
  \addConstraint{A_{way}D}{=P}{}
  \addConstraint{\sum_{t_{disc}=1}^{\frac{t_h}{\tau}}\mathcal{H}(t_{disc}) \mathcal{T} }{D \leq h}{}
  \addConstraint{\sum_{i=1}^n \underline{d} \leq  A_{dyn}D}{\leq \bar{d}}{}
  \label{trajectory NLP}
\end{mini}

where $H_{net}$ is formulated by $H(\text{Obs}) + H(\text{Fin})+H(t+t_h)$ and $F_{net}$ by $F(\text{Fin})+ F(\text{Obs})$

The Non Linear Program in Eq. \eqref{trajectory NLP} is a Convex QP and can be solved using available solvers.

To specify the end time if the final pose's desired time is unavailable, We utilize newton's second law of motion by taking the magnitude(Euclidean Norm) of the position,velocity and acceleration. This while neglecting interactions, allows for a fixed time to be specified beforehand, thereby simplifying the optimization problem. Moreover, the re-evaluation of the end time during every re-planning allows it to be much more flexible. In the future we will look to incorporating optimization of time into the problem.    

The implementation of algorithm using B-splines is different when there are non-zero/free end states as B-splines' end derivatives are zero and require iterative optimization for non-zero derivatives \cite{de1978practical}. To overcome that problem, our attempt is by extending the time horizon appropriately so as to mitigate the end derivative as zero, that is we attempt to ensure that the known desired pose doesn't come at the end of the B spline.

To handle infeasible QP that arise due to the inequality constraints and number of constraints. We utilize a two step process for the same. In the first step, we apply the previous solution. In the second step, solve the QP again but relax the dynamic constraints to be applied only at the transition points between the polynomial splines.    

\section{Results and Discussion}
\label{Results}
The proposed algorithm was implemented in C++ and integrated into Robot Operating System(ROS). In our implementation a degree of $n$+1 was utilized with $t_{segment}$ being one second. We utilized a $\tau$ of 0.1 seconds for the other agent's prediction.The algorithm was run at a frequency of 25Hz. We tested the proposed algorithm with qpOASES\cite{qpoases} for solving the QP. 
The algorithm was tested on two different systems, a workstation with Intel Xeon E5 1630v5  processor, 32GB of RAM and a Nvidia Quadro M4000 GPU and a laptop with Intel i7-6700U Processor with 8GB of RAM and Nvidia GeForce GTX 960M GPU.  

\begin{figure}
\subfloat[][]{\includegraphics[width=0.23\textwidth]{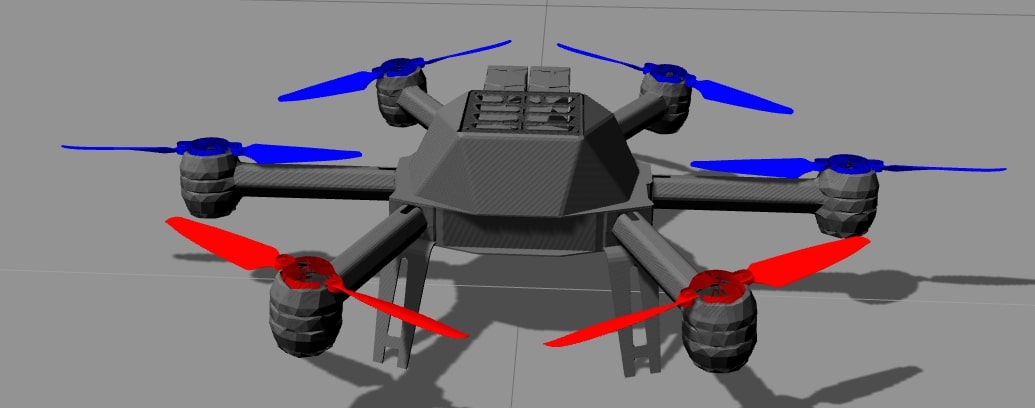}\label{Neo}} 
\hspace{4mm}\subfloat[][]{\includegraphics[width=0.20\textwidth]{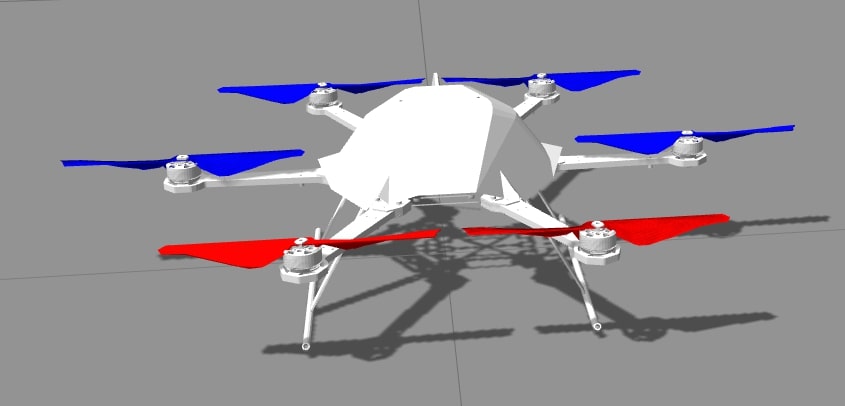}\label{Firefly}}
\caption{The aerial robots used in the simulation experiments.Fig \ref{Neo} shows an AscTec Neo with 11 inch propellers and Fig \ref{Firefly} shows AscTec Firefly}
\label{robots}
\end{figure}

The proposed algorithm was tested with two different sets of robots:
\begin{enumerate}
\item Turtlebot 3 Burger and Waffle, a second order differentially driven system utilizing the native sensor suite available on these robots. To keep the simulations realistic, the sensor update rates were unaltered. The generated trajectory is tracked using the Linear MPC proposed in \cite{mpc} which was solved using qpOASES \cite{qpoases} and had a time horizon of 2 secs with discretization of 10Hz. The trajectories were planned for a second order system. 

\item Aerial robots, a fourth order differentially flat system \cite{mellinger2011minimum}, using a rotary aerial vehicle simulator, RotorS \cite{rotors}. Furthermore, the aerial robots were mounted with a Velodyne Puck but the 3D sensor was modified to detect measurements in 2D and restricted its range to 5m to test the performance better. Moreover, the sensor's physical weight was reduced as AscTec Firefly did not have the capability to produce sufficient thrust to move the LiDAR natively. The generated trajectory was tracked using \cite{taecontroller}, a geometric controller that has been shown to be versatile and simple computationally. The robot's yaw is kept free as translation dynamics is not affected by yaw. For the experiments the altitude of the Fireflys were fixed at 1m but the altitude of the Neos was kept at 1.75 as the Neo if spawned at a height of 1 was lower than the Firefly due to the different center of masses. The robots in gazebo is shown in Fig. \ref{robots} 
\end{enumerate}

The experiments included both homogeneous and heterogeneous interactions in different environments like obstacle-free environment, intersection-like environment and unstructured environment. A depiction of the robots traversing though unknown environments is shown in Fig \ref{image_robots}. The robots were initialized randomly by generating three random number two of which was for defining position with a physical constraint of it being between -10 to +10 and orientation constrained between 0 to 360 as a whole number.
\subsection{Obstacle-free environment}
The algorithm was tested in an environment without any obstacles and with 2-10 Turtlebots. The robots were able to navigate in a collision-free manner. We also extended the test by increasing the density of the robots by having the robots concentrated within a smaller area. The test also included spawning both the variety of robots hence making it heterogeneous. The robots were able to avoid collisions amongst one other and showed good usage of free space.

\begin{figure*}
\subfloat[][]{\includegraphics[width=0.5\textwidth]{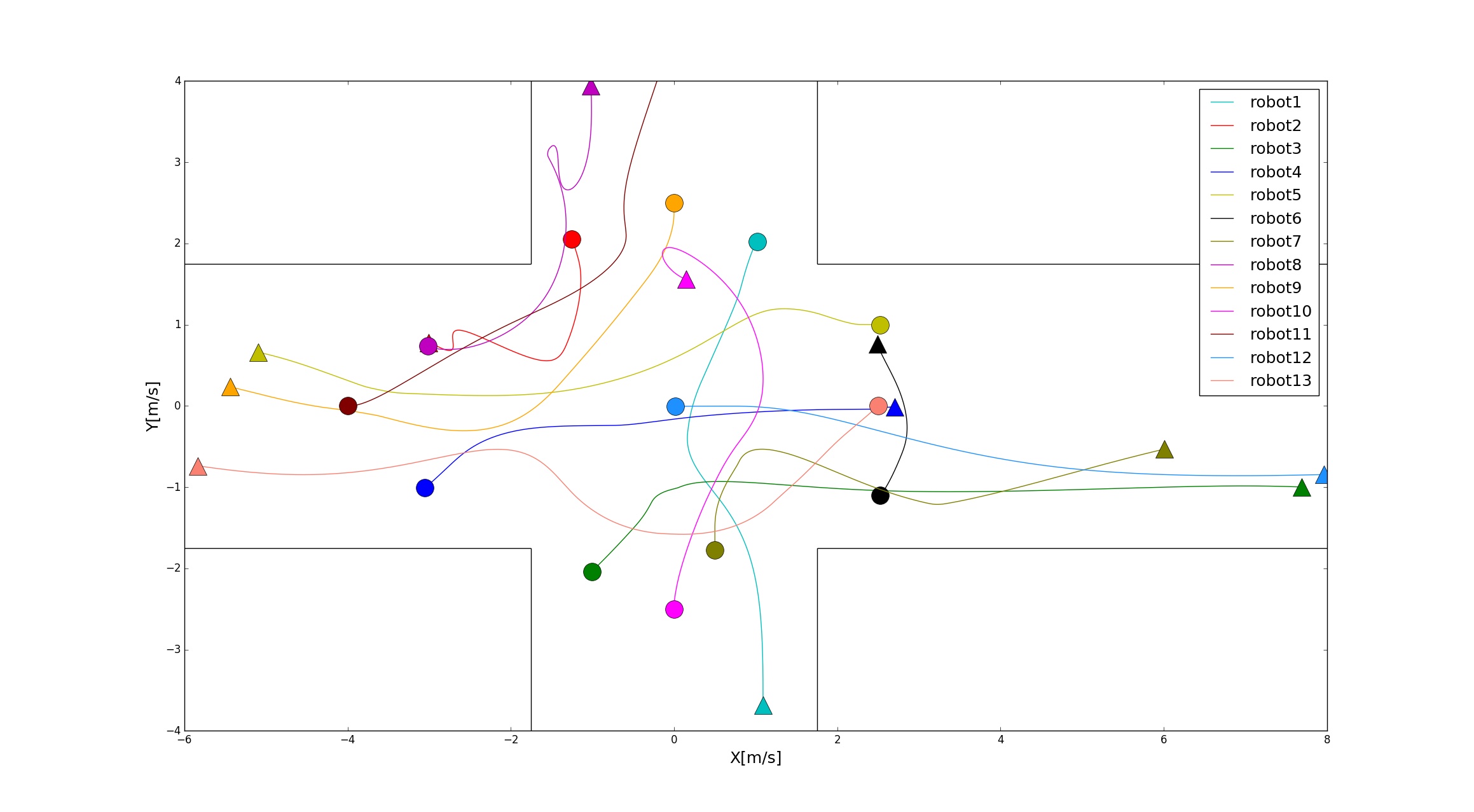}\label{13_traj}} 
\subfloat[][]{\includegraphics[width=0.5\textwidth]{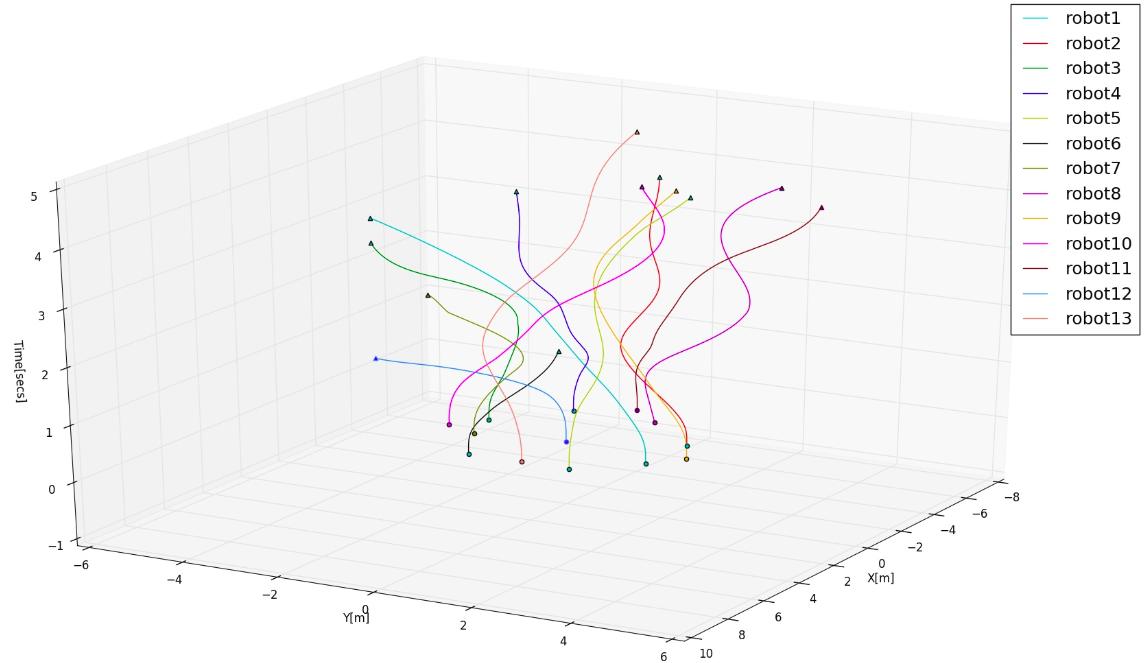}\label{13_time}}
\caption{The trajectories for thirteen aerial robots(Seven AscTec Fireflys and Six AscTec Neos) traversing across intersection-like environment. Fig \ref{13_traj} shows the trajectories in the two dimensional environment with the intersection bordered as gray walls. Fig \ref{13_time} shows the trajectories for the robots with time as the third axis. Colored circles represent the starting point of the trajectories and the colored triangles represent the end pose of the trajectories}
\label{thir_traj_inter}
\end{figure*}

\subsection{Intersection-like Environment}
A cross road-like structure was formed with the help of walls that resembled an intersection. We utilize an intersection-like environment as it is an important usage of the labeled multi-robot problem. This environment is also a good example to test the algorithm on as the walls are unknown obstacles for the robots and constrain the available free space. The robots were able to avoid collisions in most of the scenarios but showed collisions at certain points and/or veered off trajectories too.

In the first set of tests, using Turtlebots, the robots were placed randomly but with minimal interactions or cluttering of the space as the robots were comparatively smaller. The algorithm was tested with upto ten robots in different configuration(number of waffles and Burgers). The robots were also given non-zero(free) end poses so as to illustrate the functioning of an intersection.

\begin{figure*}
\subfloat[][]{\includegraphics[width=0.5\textwidth]{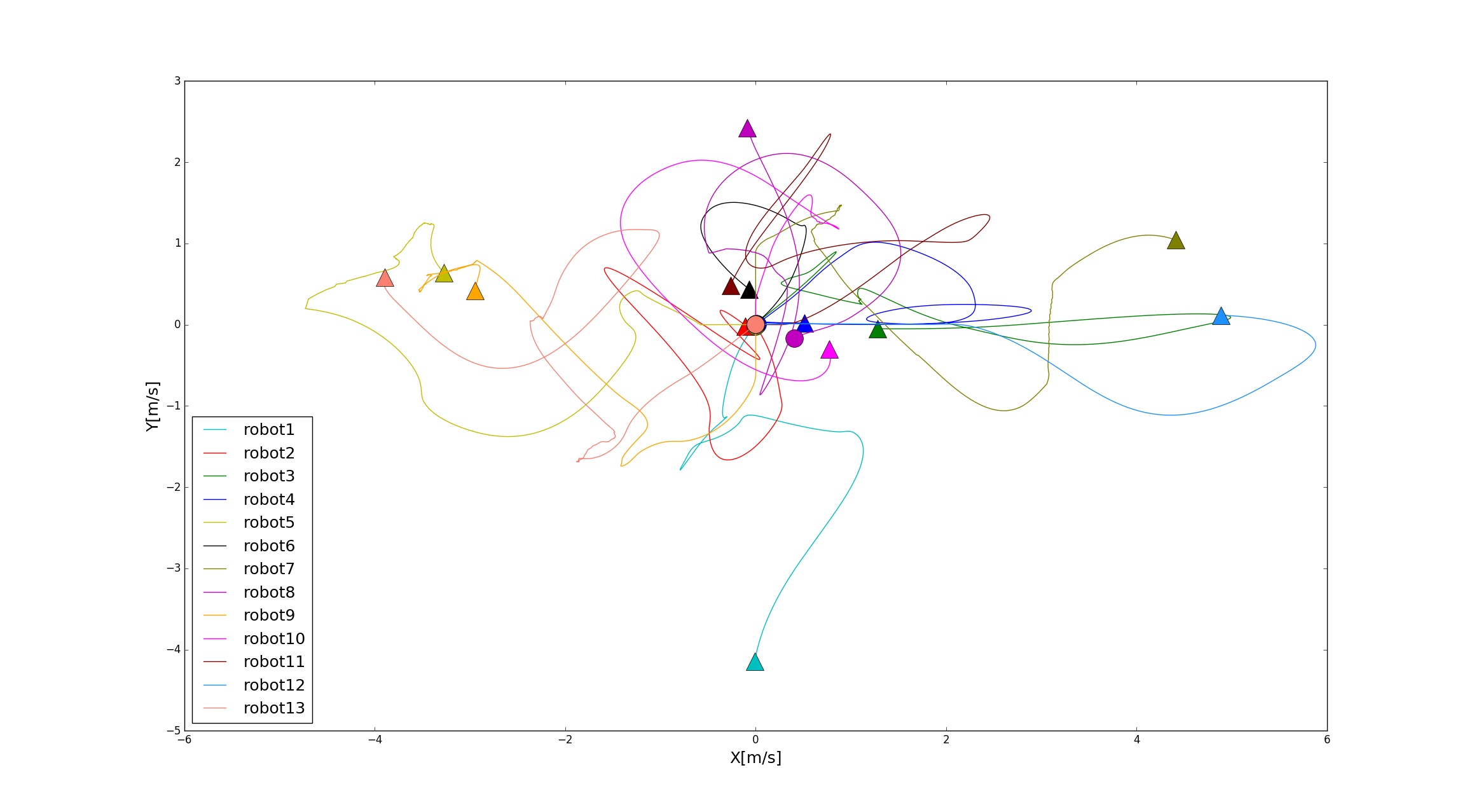}
\label{13_velover}} 
\subfloat[][]{\includegraphics[width=0.5\textwidth]{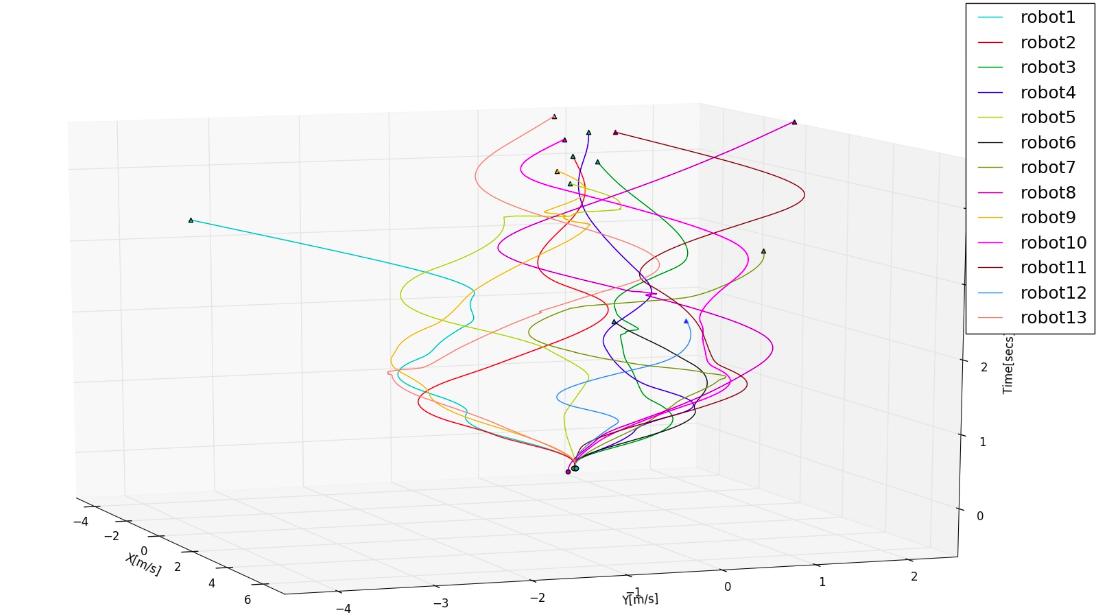}
\label{13_vel3d}} 

\subfloat[][]{\includegraphics[width=0.5\textwidth]{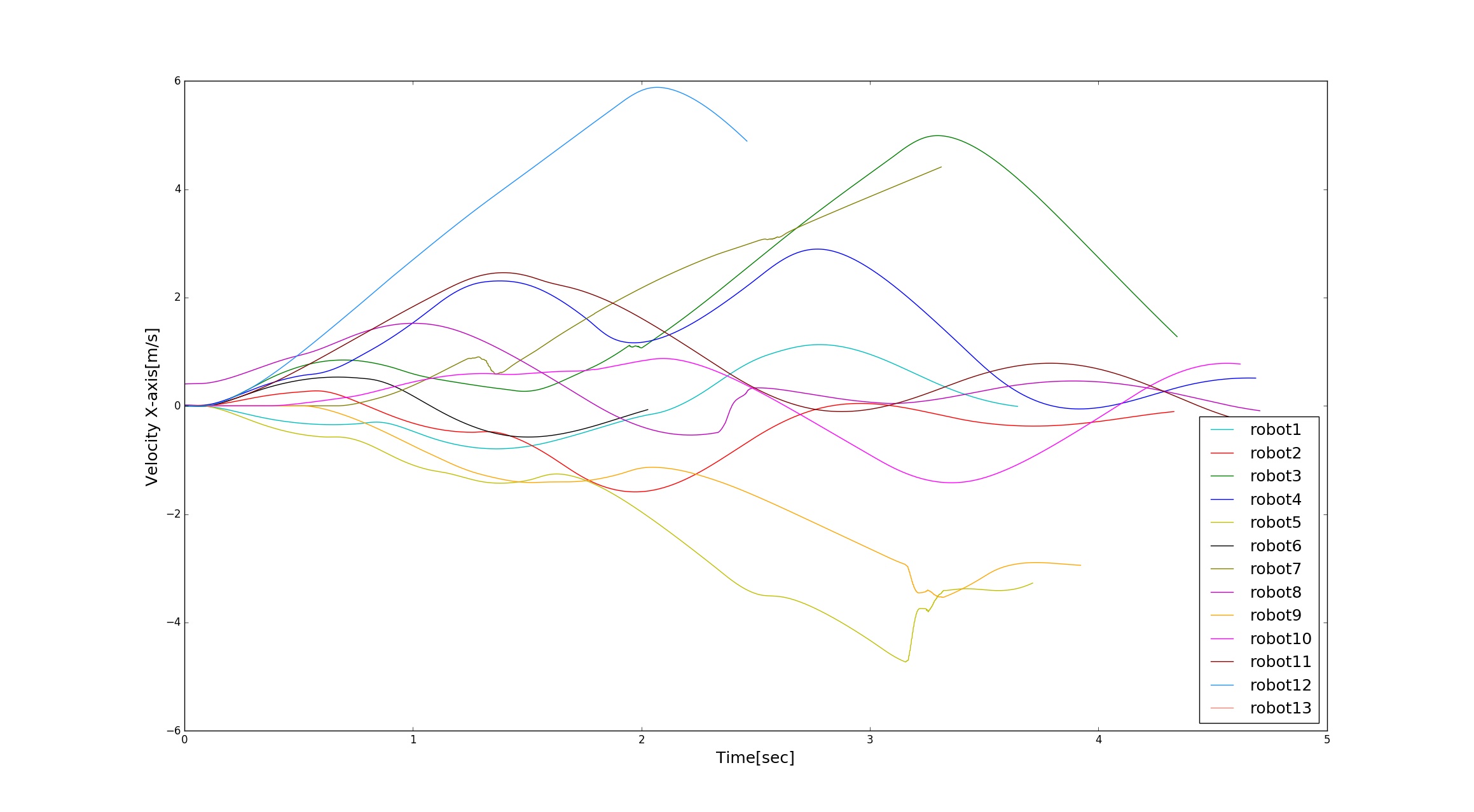}\label{x_inter}}
\subfloat[][]{\includegraphics[width=0.5\textwidth]{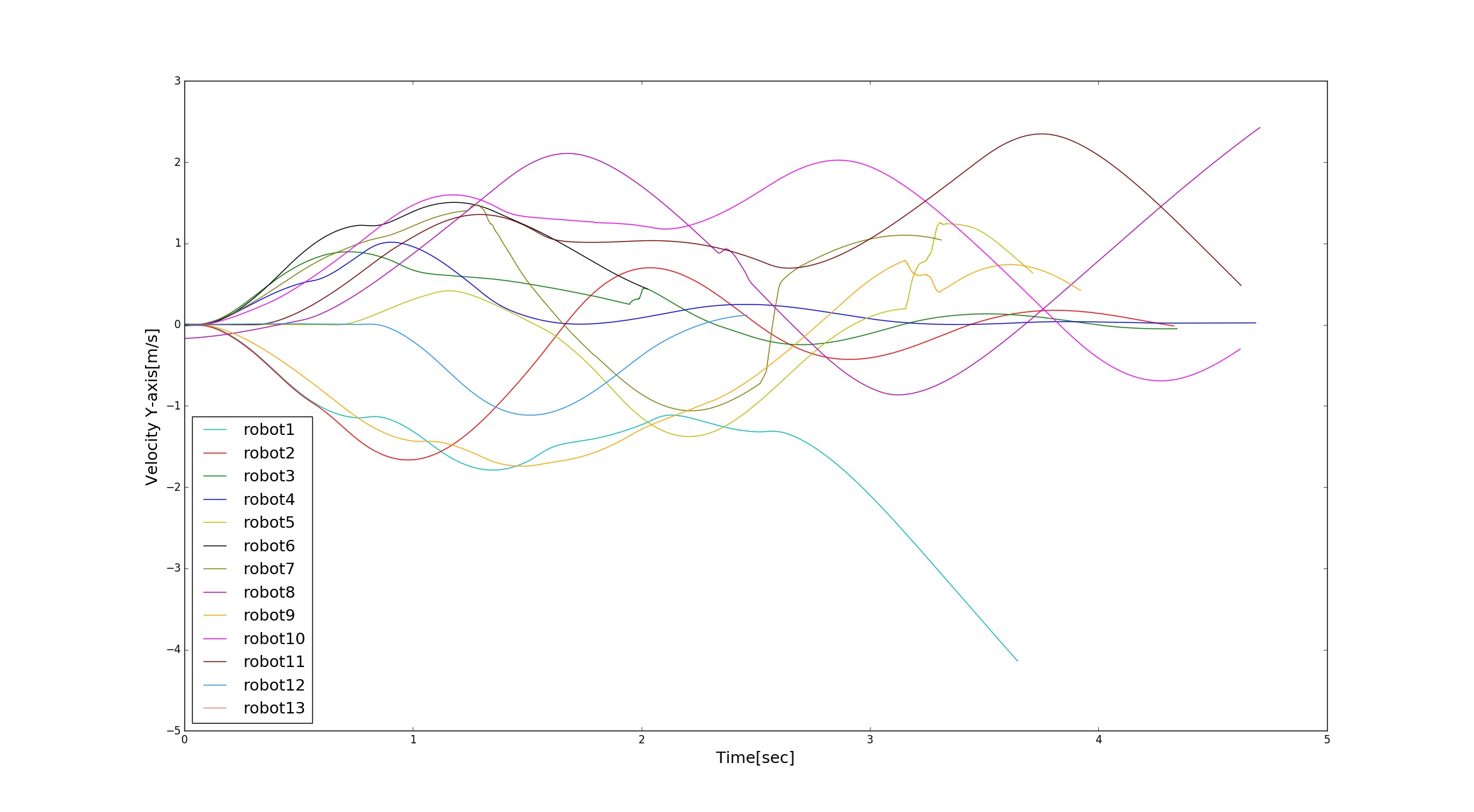}\label{y_inter}}
\caption{The velocity profiles for the robots as they traverse across the intersection. Fig \ref{13_velover} shows the x and y velocity trajectory while Fig \ref{x_inter} and \ref{y_inter} shows the velocity trajectory through time in each axes individually}
\label{Vel_inter}
\end{figure*}

In a second set of experiments with aerial robots,to test the capabilities with higher order robots, thirteen aerial robots(Six AscTec Neos and Seven AscTec Fireflys) were tested with the robots starting off near the center of the intersection. The robots were given arbitrary end poses with few robots given non-zero end velocities.  The trajectories for the robots are shown in Fig \ref{thir_traj_inter}. Fig \ref{13_time} shows the trajectories of the robots with time as the third axis to showcase the collision avoidance maneuver. The trajectories generated are smooth and the robots evade collisions smoothly. From Fig \ref{13_time}, it can inferred that the robots are able to avoid collisions in congested areas with smooth maneuvers. Fig \ref{Vel_inter} shows the velocity trajectories as the robot traverse across the intersection. The velocity trajectory generated is smooth, and doesn't show much rapid changes. But the velocity is comparably slow and constantly changing as the robots evade across each other in cluttered areas but as the free space increases the robot accelerate to higher speeds.

\subsection{Unstructured Environment}
The algorithm was tested in an environment with randomly spaced obstacles of different sizes and shapes. 

In the first ground robot based tests, heterogeneous robots consisting of upto four waffles and three burgers in an environment spawned sparsely but in close proximity to obstacles. The robots were given non-zero end desired poses. The robots were able to formulate trajectories that enabled them to avoid collisions with other robots and obstacles but had a tendency to collide with irregular shaped obstacles in the environment. This is not surprising as the robots only have 2D map of the environment and the third dimension still exists.

\begin{figure*}
\subfloat[][]{\includegraphics[width=0.18\textwidth]{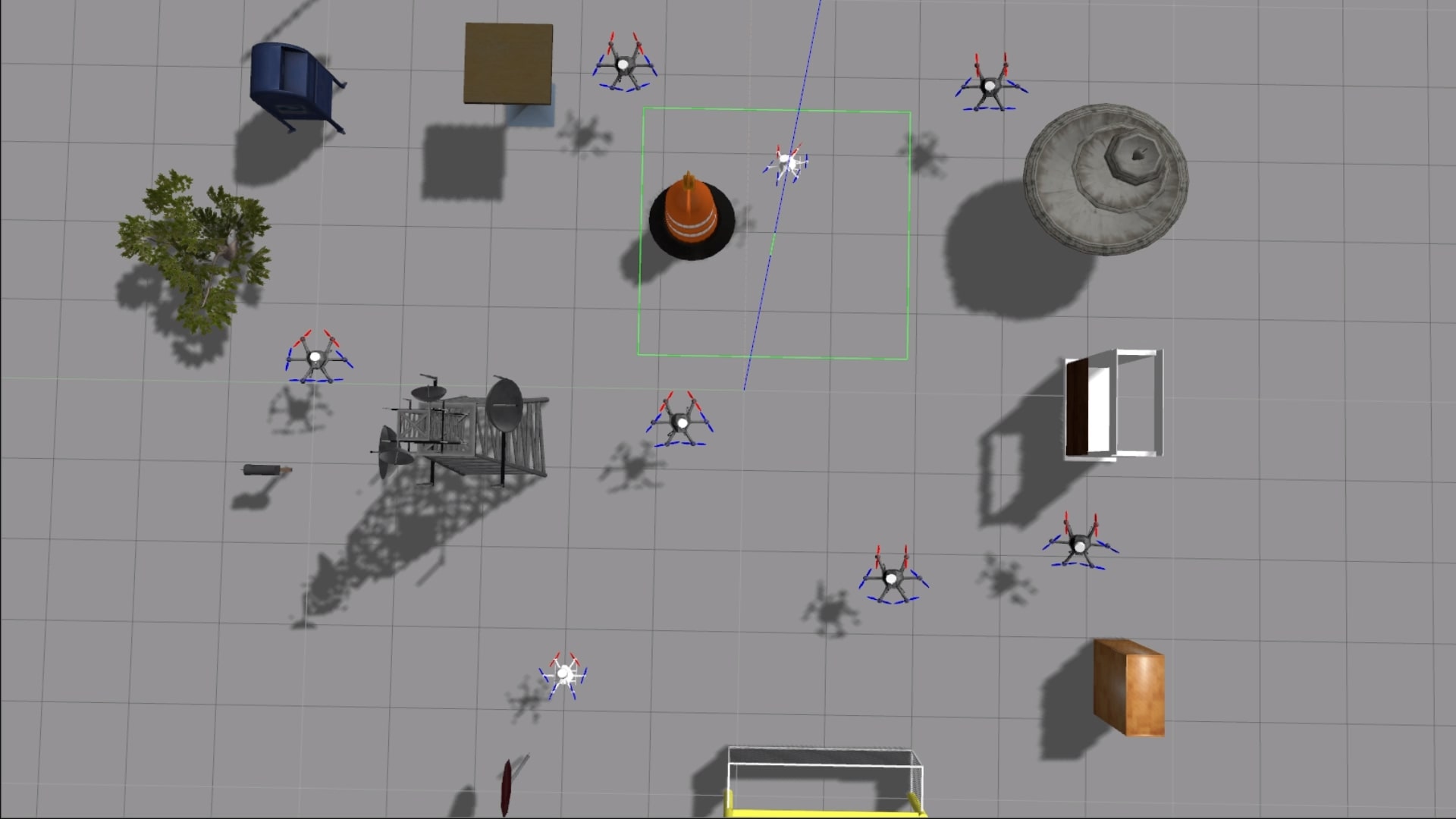}
\label{Unstructured_aerial1}} 
\hspace{2mm}
\subfloat[][]{\includegraphics[width=0.18\textwidth]{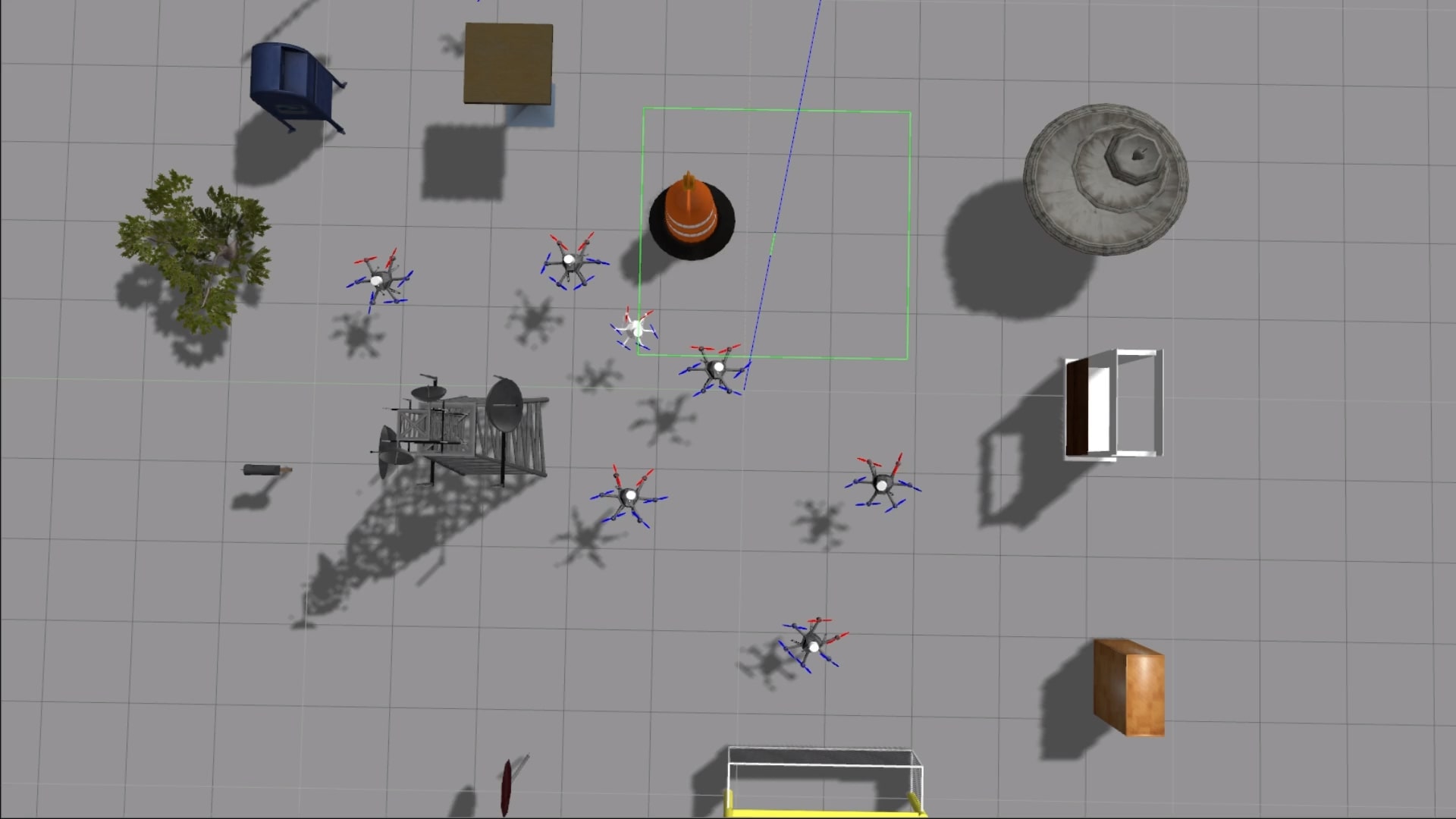}\label{Unstructured_aerial2}}
\hspace{2mm}
\subfloat[][]{\includegraphics[width=0.18\textwidth]{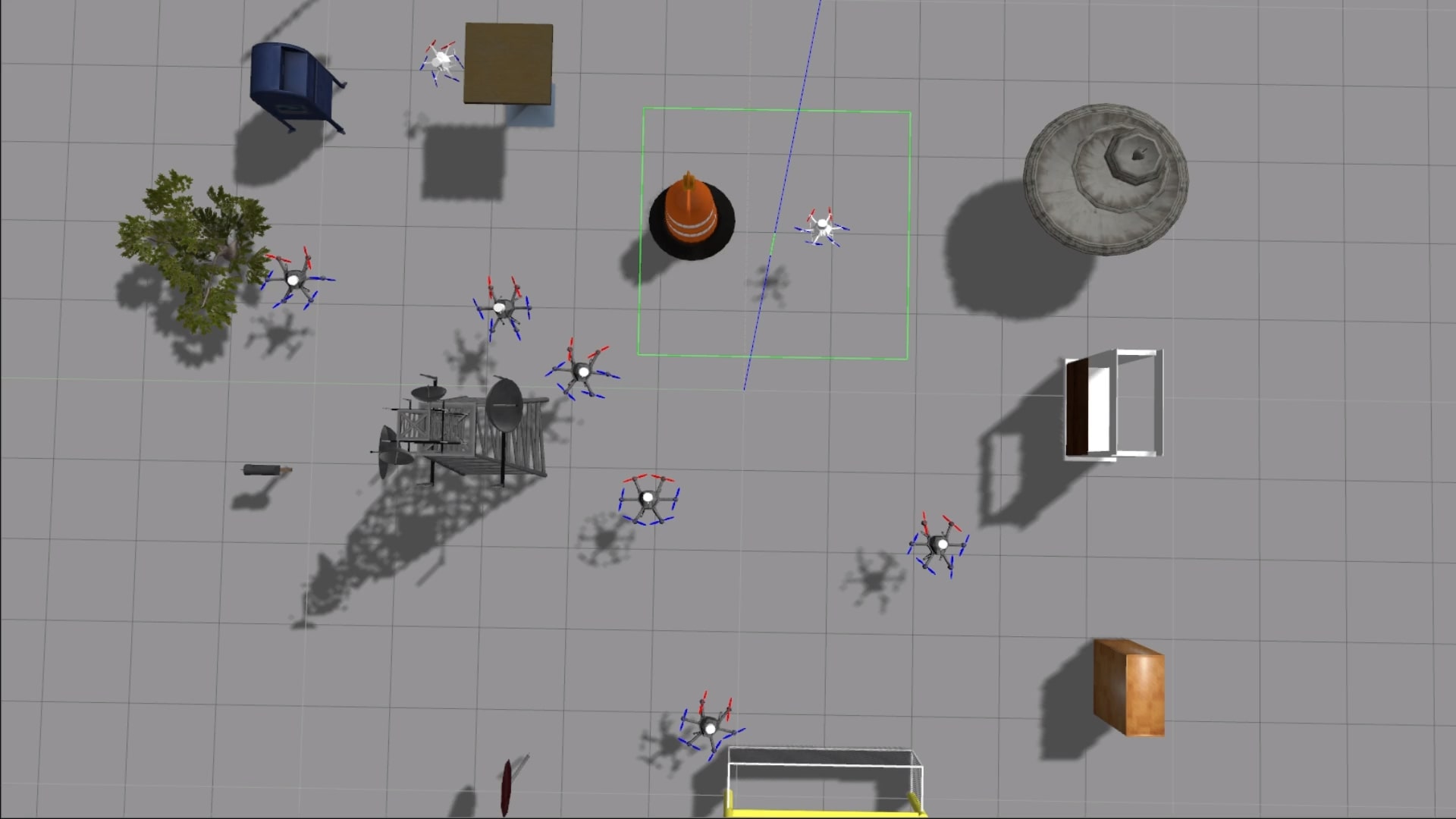}\label{Unstruct3}}
\hspace{2mm}
\subfloat[][]{\includegraphics[width=0.18\textwidth]{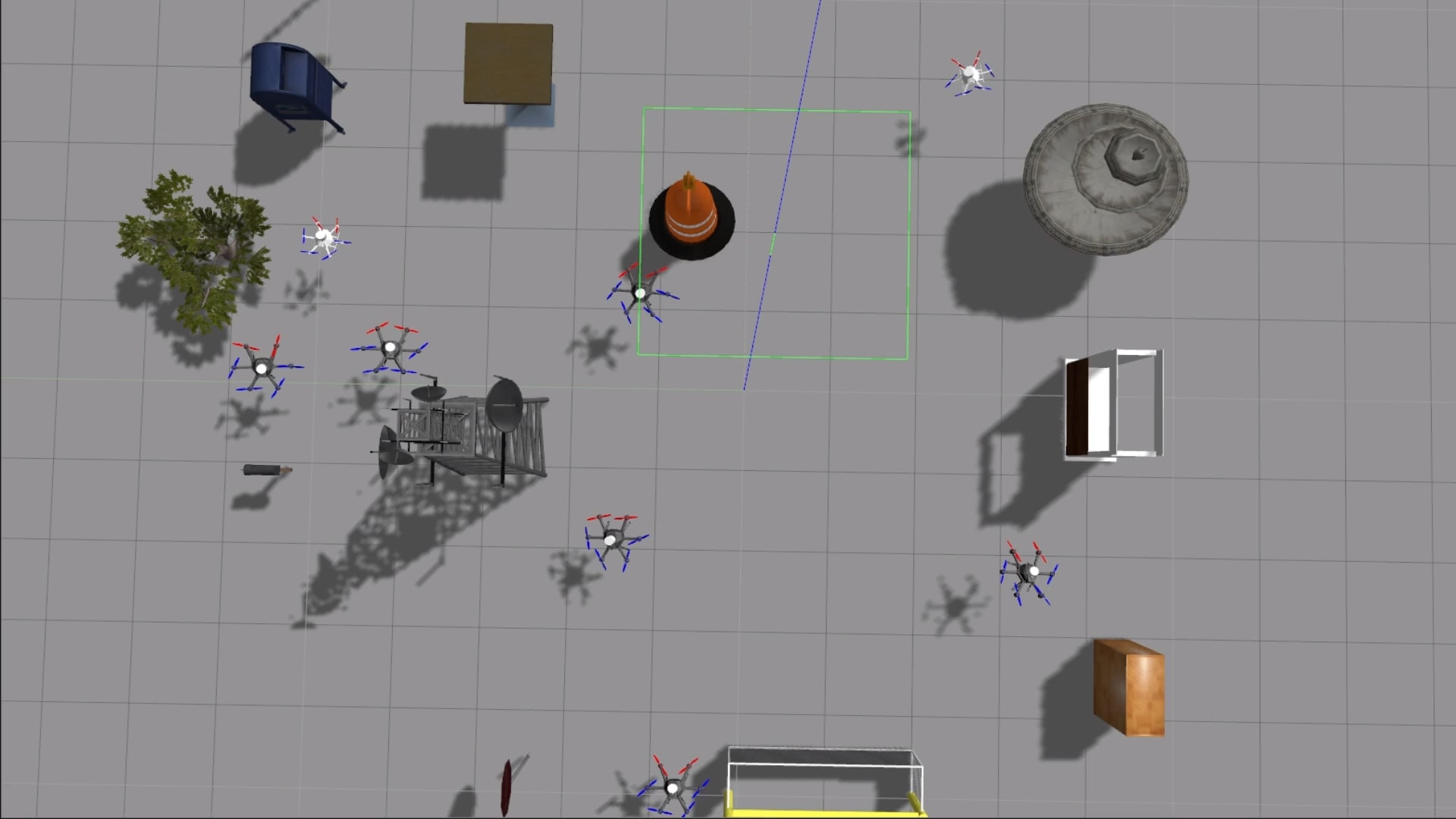}
\label{Unstructured_aerial4}} 
\hspace{2mm}
\subfloat[][]{\includegraphics[width=0.18\textwidth]{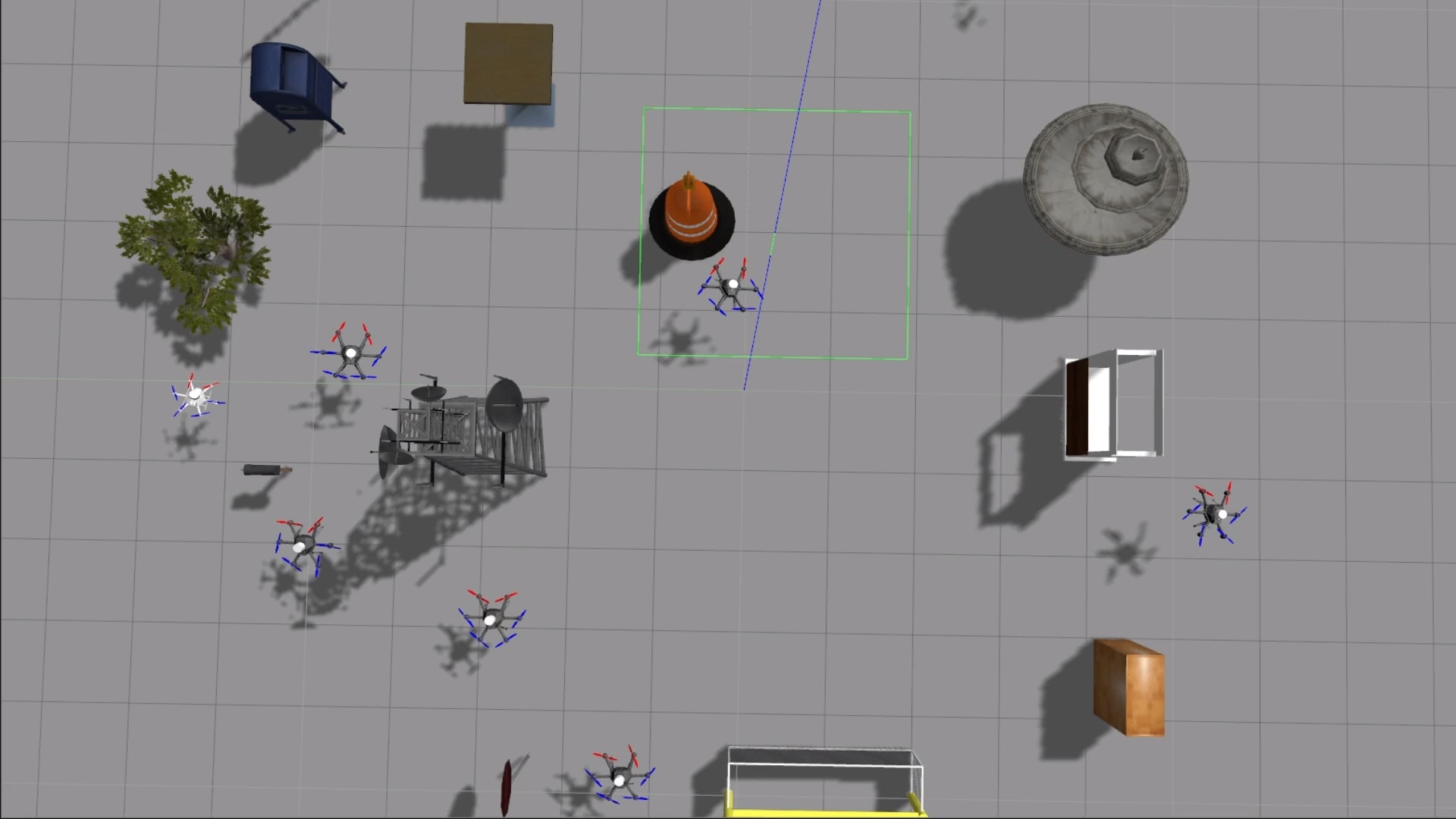}\label{Unstructured_aerial5}}

\subfloat[][]{\includegraphics[width=0.18\textwidth]{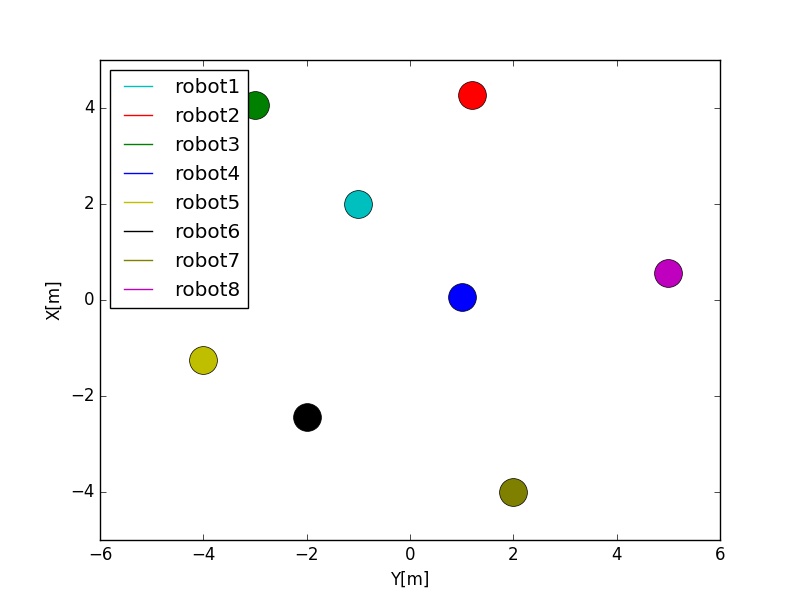}
\label{unstructure_aerial1}} 
\hspace{2mm}
\subfloat[][]{\includegraphics[width=0.18\textwidth]{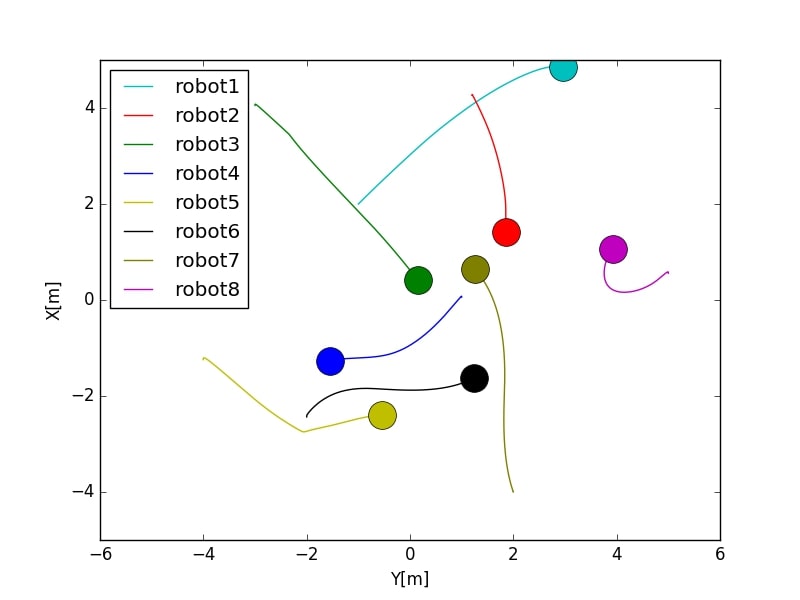}\label{unstructured_aerial2}}
\hspace{2mm}
\subfloat[][]{\includegraphics[width=0.18\textwidth]{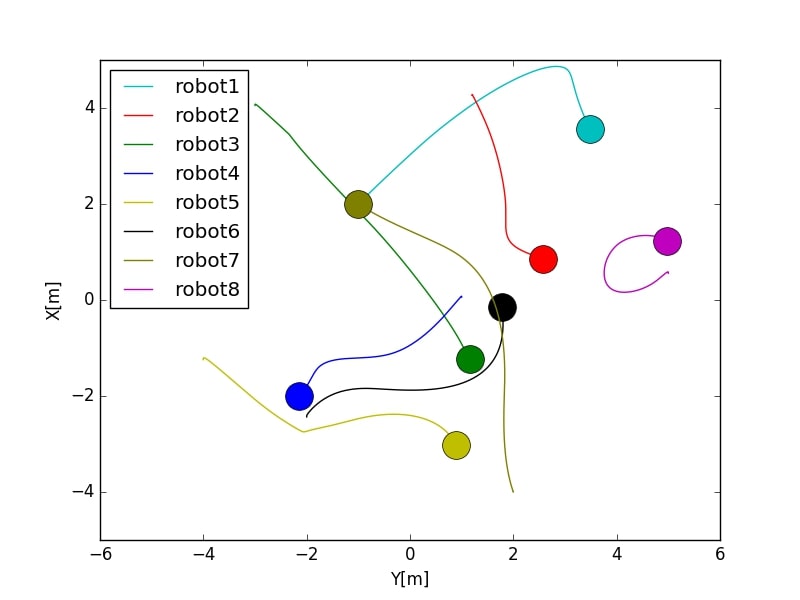}\label{unstruct3}}
\hspace{2mm}
\subfloat[][]{\includegraphics[width=0.18\textwidth]{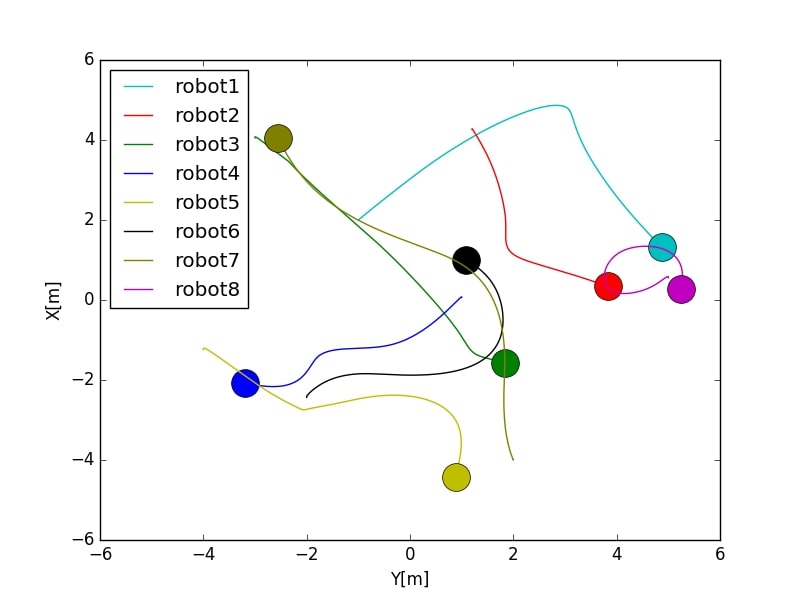}
\label{unstructured_aerial4}} 
\hspace{2mm}
\subfloat[][]{\includegraphics[width=0.18\textwidth]{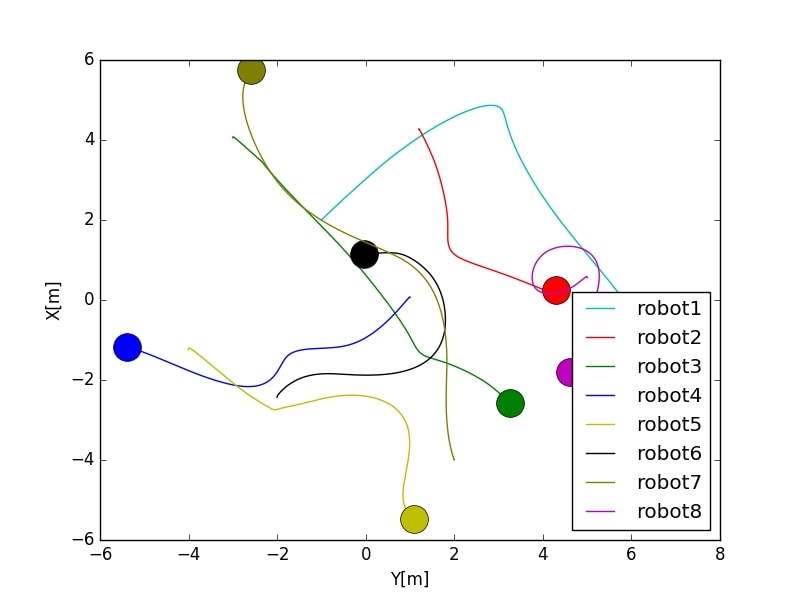}\label{unstructured_aerial5}}
\caption{Snapshots of robots and the trajectories they have completed while eight aerial robots are progressing across an unknown unstructured environment. The colored circles represent the robots' current position at the snapshot}
\label{unstruct_aerial}
\end{figure*}

\begin{figure*}
\subfloat[][]{\includegraphics[width=0.5\textwidth]{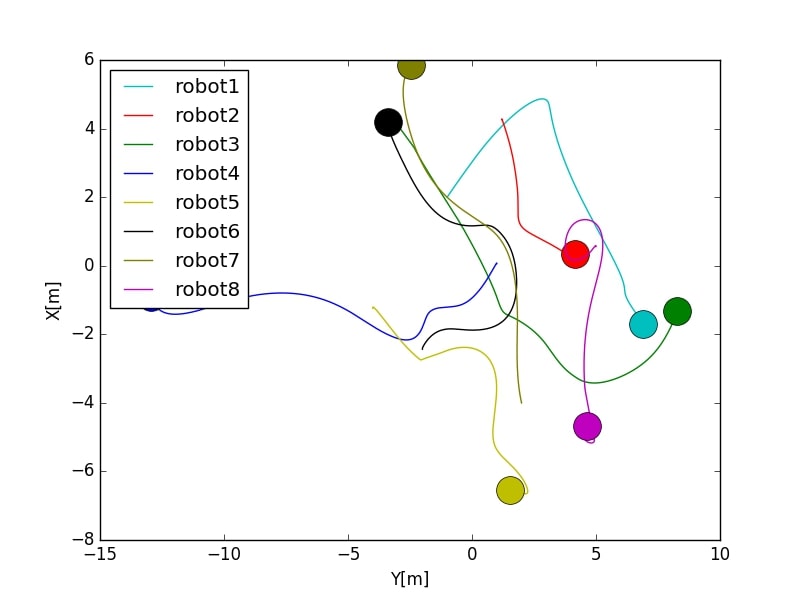}\label{traj_unstruct}} 
\subfloat[][]{\includegraphics[width=0.5\textwidth]{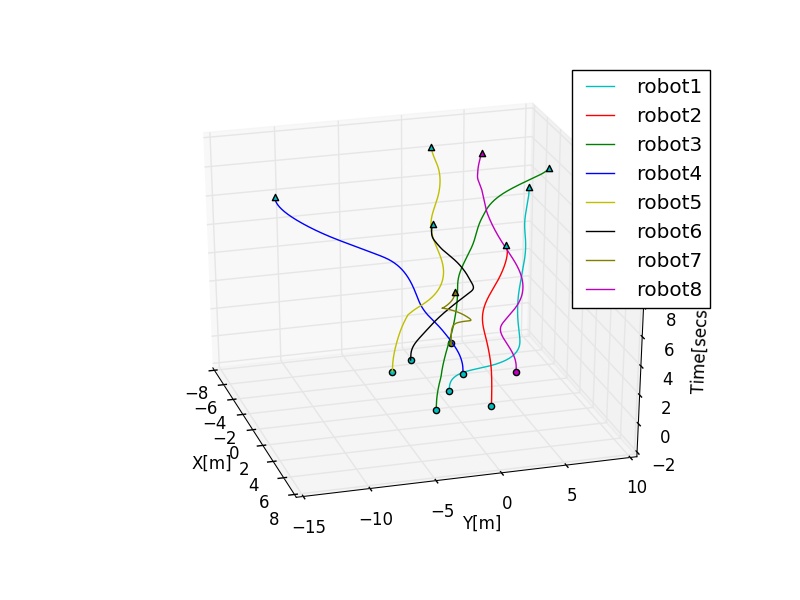}\label{unstruct_time}}
\caption{The trajectories as Eight aerial robots(Two AscTec Fireflys and Six AscTec Neos) traverse across the unstructured environment. Fig \ref{traj_unstruct} shows the trajectories in the two dimensional environment. Fig \ref{unstruct_time} shows the trajectories for the robots with time as the third axis. Colored circles represent the starting point for the trajectories and the colored triangles represent the end pose of the trajectories}
\label{eight_traj_unstruct}
\end{figure*}

\begin{figure*}
\subfloat[][]{\includegraphics[width=0.5\textwidth]{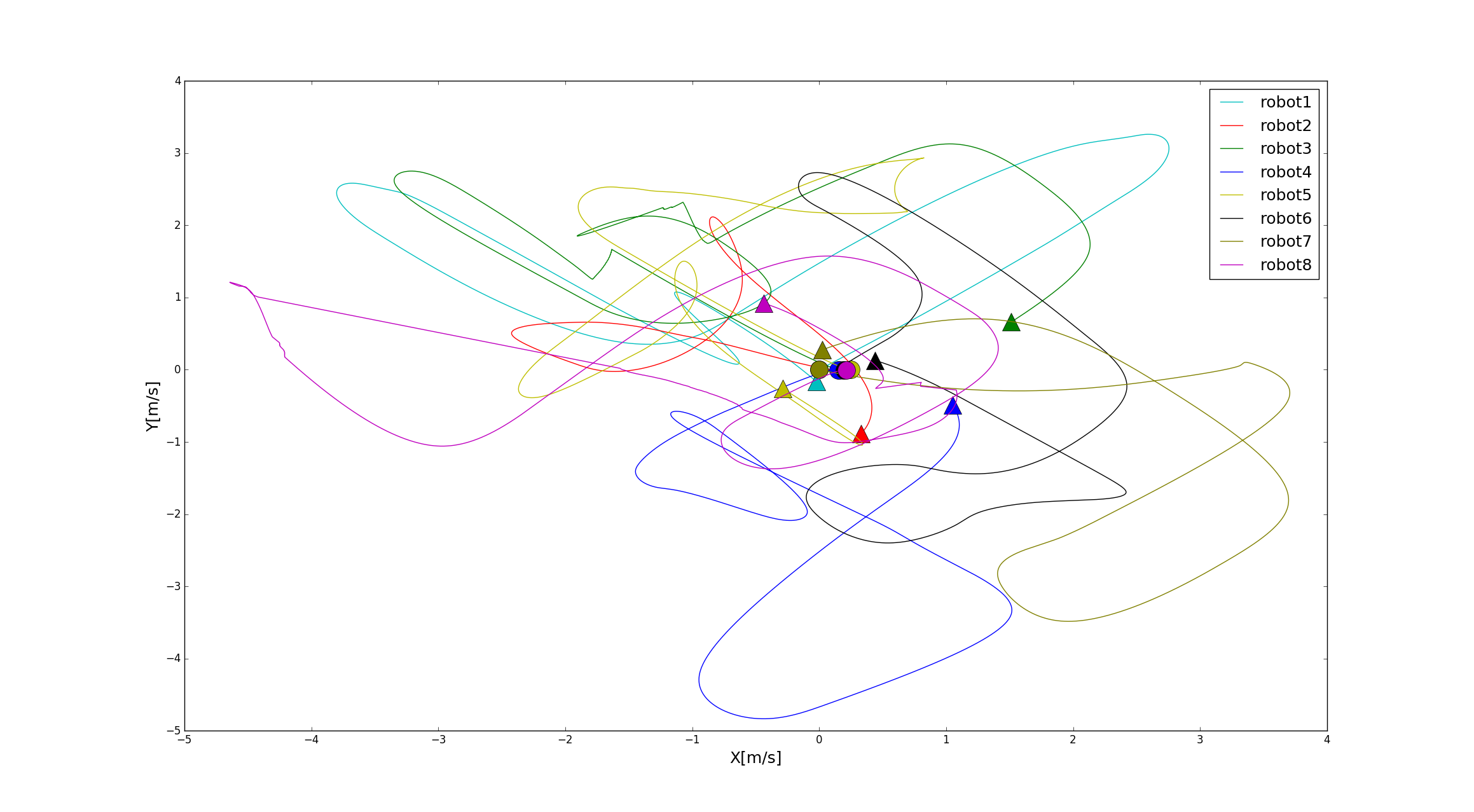}
\label{8_velover}} 
\subfloat[][]{\includegraphics[width=0.5\textwidth]{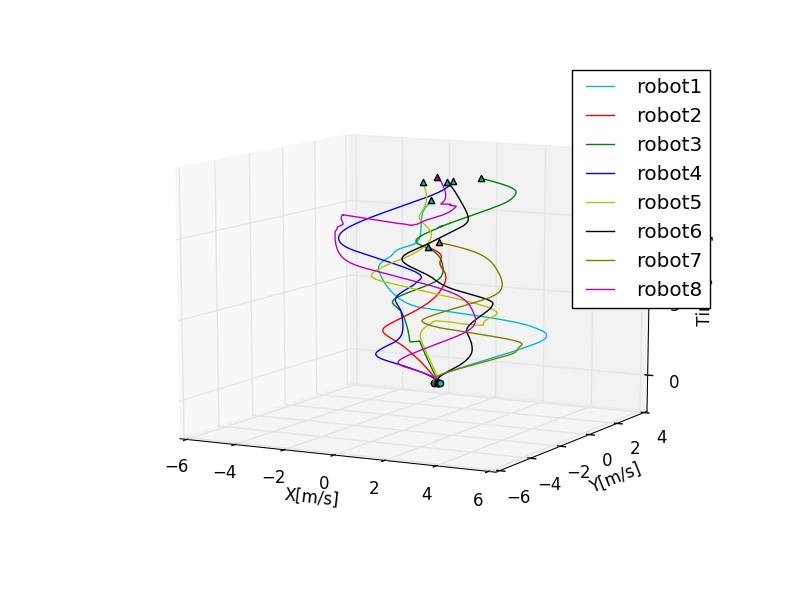}
\label{8_vel3d}} 

\subfloat[][]{\includegraphics[width=0.5\textwidth]{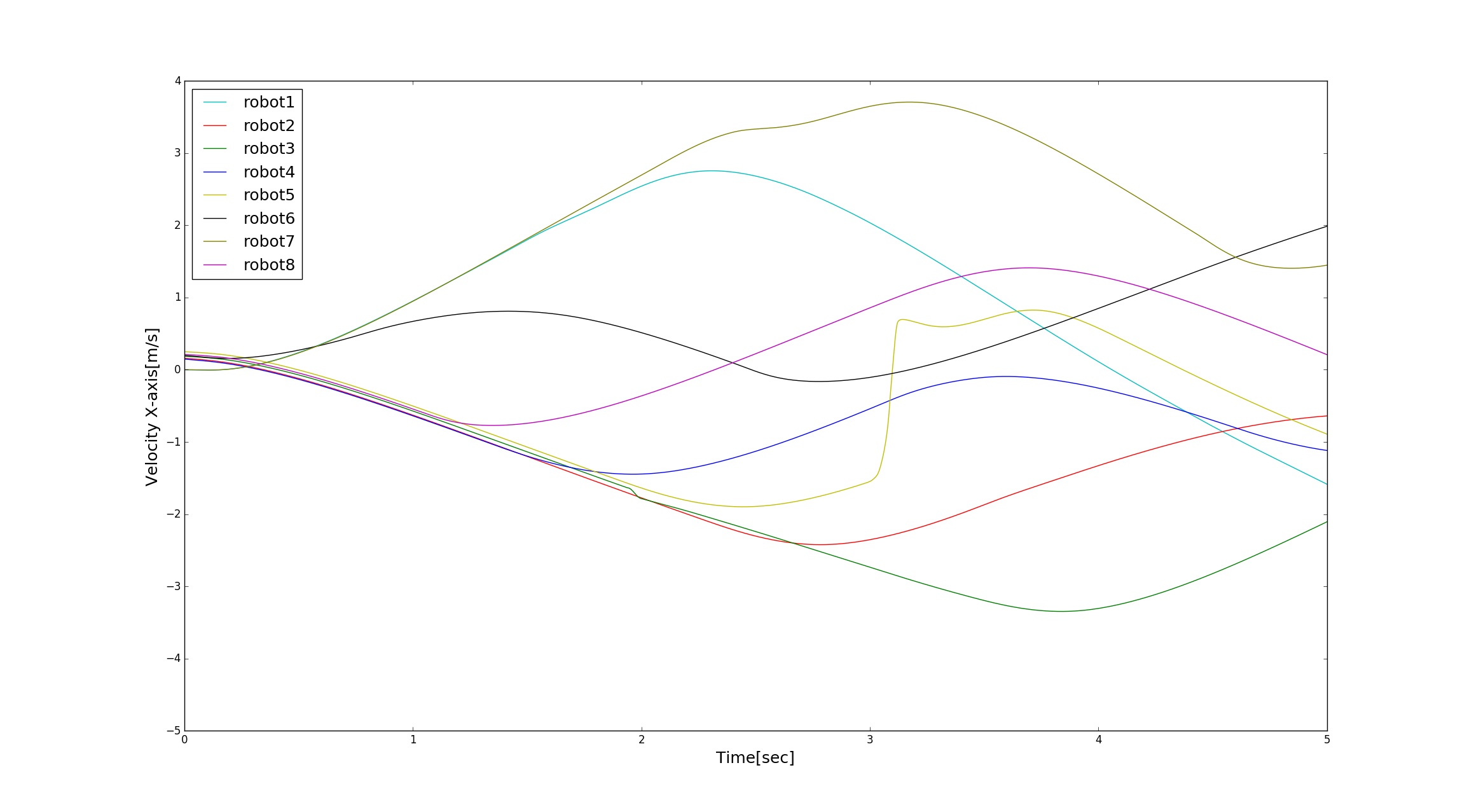}\label{x_unstruct}}
\subfloat[][]{\includegraphics[width=0.5\textwidth]{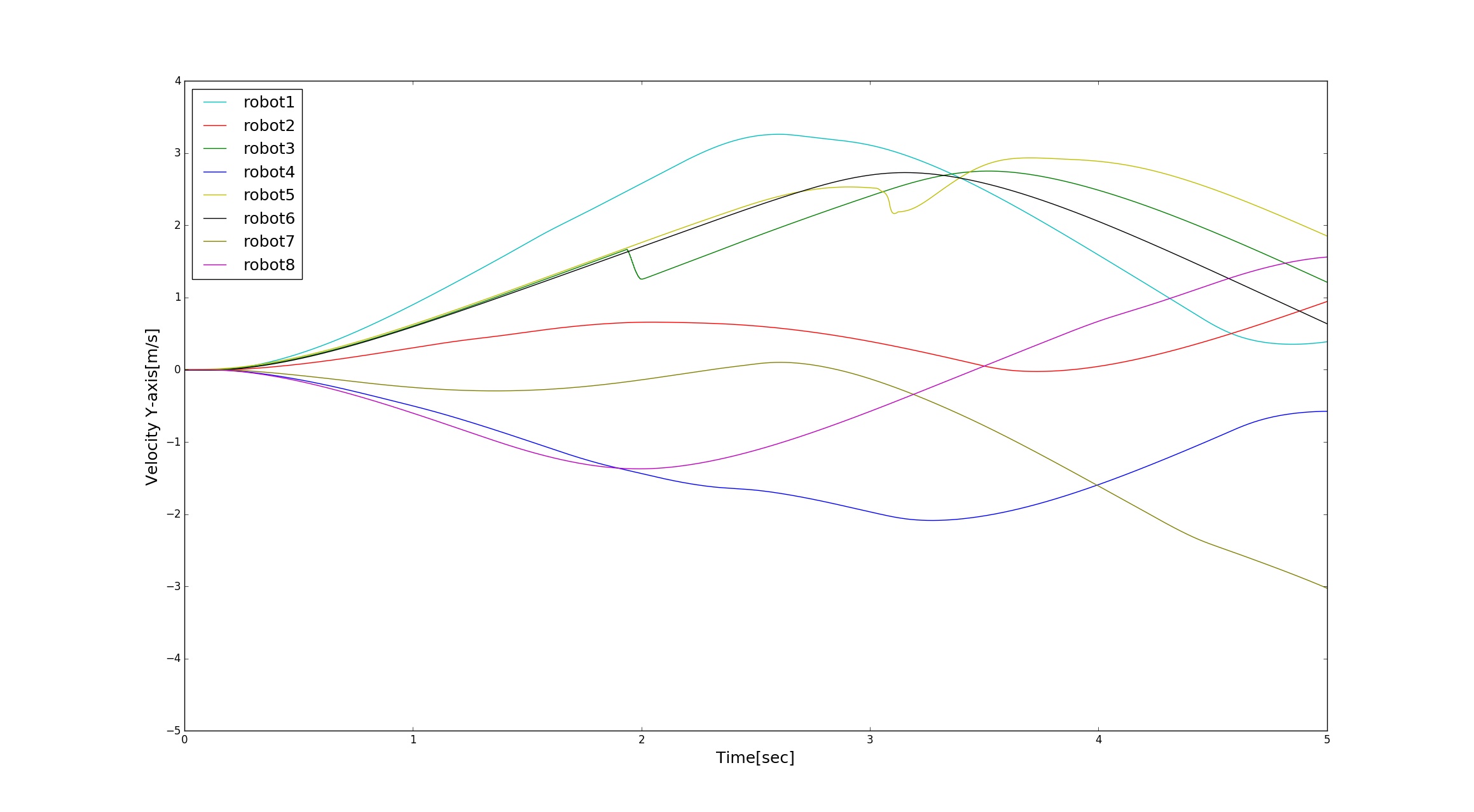}\label{y_unstruct}}
\caption{The velocity profiles for the robots as they traverse across the intersection. Fig \ref{8_velover} shows the x and y velocity trajectory while Fig \ref{x_unstruct} and \ref{y_unstruct} shows the velocity trajectory through time in each axes individually}
\label{Vel_unstruct}
\end{figure*}

The algorithm was tested with eight aerial robots(Two Fireflys and Six Neos) in an unstructured environment. The unstructured environment used for the ground robots is different from the environment used for aerial robots with small obstacles removed and some obstacles added like a radio tower near the center. Moreover, the size of one of the objects(A postbox) was kept smaller so that robots can fly above it. All the robots in this experiment had to travel through multiple waypoints which were only defined for the pose. The trajectory of the robots as they travel through the environment is shown in Fig. \ref{unstruct_aerial} as a sequence of images taken at different time points to better illustrate the robots' trajectories in comparison with the obstacles in the environment. The complete trajectory of the robots is shown in Fig \ref{eight_traj_unstruct}. Fig \ref{Vel_unstruct} shows the robot velocities through the unstructured environment. In comparison to the trajectories across the intersection, in the unstructured environment the robots showcase limited drastic changes and showcase smoother evasive maneuvers through the environment. Another point observed in the velocity profile is that the robots' velocity trajectories are more concentrated in the unstructured environment. The velocities have negative profiles for robots that are moving in the negative axes as all robots are spawned with yaw angles zero. 


      

\subsection{Discussions} 
The algorithm is able to generate smooth, collision-free trajectories for different robots ranging from second order non-holonomic robots to fourth order aerial robots of different sizes and shapes but in our experiments, collisions were not completely inevitable. Some of the inter-robot collisions were due to inaccurate predictions because of the difference in polynomial representations. While others were due to tracking errors. Moreover, in the attached video, in the intersection-like environment, it can be seen that some collisions were avoided due to high roll and pitch angles rather than the good collision-avoidance maneuvers (But these maneuvers all avoided collisions with the motor boom but would have collided at most only with propeller). Nonetheless, the algorithm shows an efficient trajectory generation. In the future we will look to add tracking error bound and/or propeller sizes also into the robot sizes, something we did not account for to increase traversable space. The collisions with obstacles were due to the inaccurate representation of the obstacles and also the unaccounted sensory noise. Moreover, the LiDAR measurements are available at 5Hz whereas the trajectory optimization algorithm runs at 25Hz. Another observation to be noted is that the collision avoidance is attempted in discrete time rather than continuous-time despite using continuous time representations. So in the future developing a method to evaluate collisions by interpolating the trajectories is important. The algorithm also scales well due to the usage of safe region that is irrespective of the number of robots and is more of a formation of safe region by an intersection of convex planes. But there have been scenarios where a great number of convex planes were created. 

The trajectory prediction algorithm despite the lack of unique identification did not mislabel the robots' states due to the usage of velocity and acceleration into the verification. But some mislabels were detected if robots were moving in close proximity in the same directions.

The algorithm, despite some hiccups and some challenges to be overcome, showcases a good performance for collision-free navigation of multi-robot. It is able to handle non-zero end velocity and fixed end time of pose (two of the rarely attempted avenues in multi-robot applications) considerably well. But we believe that for better trajectory generation, the dynamic limits have to accounted for in a better manner. That is something which the algorithm does not account for properly as the dynamics are separated per axis and used as box constraints but in reality the dynamic limits are coupled across the different dimensions often resulting in euclidean norm better describing the dynamic limits especially with acceleration of multirotors which is a non convex donut shaped constraint. 

Another point to be noted is that, the AscTec Neos showed some oscillations at the end of the trajectories in both intersection-like environments and Unstructured Environment. According to our debugging, the higher number of objects in the simulation environment results in an  increased load on the physics engine as the trajectory tracking algorithm was tuned appropriately. Moreover another possible reason is the addition of the LiDAR on the robot affecting the moment of inertia. We are actively working on understanding the reasoning for this anomaly and will rectify it in the future.  

\section{Conclusion}
\label{Conclusions}
A decentralized algorithm for collision-free trajectories of multiple robots in unknown two dimensional environments was proposed in this work.The proposed method was tested extensively in simulations using gazebo for different sets of robots(non-holonomic ground robots and underactuated aerial robots) in intersection-like and  unstructured  environments.

The algorithm parametrized trajectories by B Splines to allow for a continuous-time representation by exploiting the differential flatness property shown by a variety of robotic systems. For collision avoidance, we used a method for generating safe convex regions by using prediction of other robots' trajectories and detection of obstacles using a LiDAR sensor. A method for obstacle detection was proposed that allowed for simpler collision avoidance verification with respect to the obstacles and parameterized representations. The algorithm was broken down into smaller parts to ensure that, in a manner akin to the Real Time Iteration of nonlinear model predictive control systems, the trajectories can be generated without delay. The approach has proven to be capable of generating smooth trajectories in constrained environments like intersections. Furthermore, it also showed a good evasive performance for collision avoidance. Due to simulations, each robot utilized a single thread for its processing, a real life experimental implementation of the algorithm on multiple different processors can utilize multiple cores for the processing thereby resulting in faster run times. 

In the future there exist many different avenues to build upon the current work. The collision representation is a discrete time representation and a continuous-time representation of the collision is an avenue for future research. Such approaches can be attempted using swept volume \cite{swept} for example. The free space generation is generic enough and can be extended to 3D environments sufficiently by just adding another dimension. Besides, the addition of a method to accommodate errors in prediction and/or tracking might allow for providing theoretical guarantees on the safety of the algorithm. Furthermore, extending the method to moving obstacles that do not transmit data is another enticing point to look at in the future.  The sensors used had low update rates. Utilizing faster sensors or utilizing RGB-D or Stereo cameras for local maps is an another important possible extension. Moreover, incorporating a higher variety of primitives for the obstacles will allow for a much more accurate obstacle representation. An extension of the proposed mapping mechanism to 3D using methods like constructive solid geometry  \cite{csg} is another avenue for future research.

\section*{Supplementary Material}
A video showcasing the experiments in Gazebo can be found on \href{https://www.youtube.com/watch?v=TVnFA73Idco&feature=youtu.be}{https://www.youtube.com/watch?v=TVnFA73Idco}




\bibliographystyle{IEEEtran}
\bibliography{references,refer}

\begin{thebibliography}{10}
\providecommand{\url}[1]{#1}
\csname url@samestyle\endcsname
\providecommand{\newblock}{\relax}
\providecommand{\bibinfo}[2]{#2}
\providecommand{\BIBentrySTDinterwordspacing}{\spaceskip=0pt\relax}
\providecommand{\BIBentryALTinterwordstretchfactor}{4}
\providecommand{\BIBentryALTinterwordspacing}{\spaceskip=\fontdimen2\font plus
\BIBentryALTinterwordstretchfactor\fontdimen3\font minus
  \fontdimen4\font\relax}
\providecommand{\BIBforeignlanguage}[2]{{%
\expandafter\ifx\csname l@#1\endcsname\relax
\typeout{** WARNING: IEEEtran.bst: No hyphenation pattern has been}%
\typeout{** loaded for the language `#1'. Using the pattern for}%
\typeout{** the default language instead.}%
\else
\language=\csname l@#1\endcsname
\fi
#2}}
\providecommand{\BIBdecl}{\relax}
\BIBdecl

\bibitem{mora2017multi}
J.~Alonso-Mora, S.~Baker, and D.~Rus, ``Multi-robot formation control and
  object transport in dynamic environments via constrained optimization,''
  \emph{The International Journal of Robotics Research}, vol.~36, no.~9, pp.
  1000--1021, 2017.

\bibitem{intersection}
L.~Riegger, M.~Carlander, N.~Lidander, N.~Murgovski, and J.~Sj{\"o}berg,
  ``Centralized mpc for autonomous intersection crossing,'' in
  \emph{Intelligent Transportation Systems (ITSC), 2016 IEEE 19th International
  Conference on}.\hskip 1em plus 0.5em minus 0.4em\relax IEEE, 2016, pp.
  1372--1377.

\bibitem{scholleing2012}
A.~Sch{\"o}llig, F.~Augugliaro, and R.~D'Andrea, ``A platform for dance
  performances with multiple quadrocopters,'' \emph{Improving Tracking
  Performance by Learning from Past Data}, vol. 147, 2012.

\bibitem{Regev2016icra}
T.~Regev and V.~Indelman, ``Decentralized multi-robot belief space planning in
  unknown environments via identification and efficient re-evaluation of
  impacted paths,'' \emph{Autonomous Robots}, vol.~42, no.~4, pp. 691--713,
  2018.

\bibitem{turpin2014concurrent}
M.~Turpin, N.~Michael, and V.~Kumar, ``Capt: Concurrent assignment and planning
  of trajectories for multiple robots,'' \emph{The International Journal of
  Robotics Research}, vol.~33, no.~1, pp. 98--112, 2014.

\bibitem{tang2018hold}
S.~Tang, J.~Thomas, and V.~Kumar, ``Hold or take optimal plan (hoop): A
  quadratic programming approach to multi-robot trajectory generation,''
  \emph{The International Journal of Robotics Research}, p. 0278364917741532,
  2018.

\bibitem{Zhou2017real}
Y.~Zhou, L.~HU~HS \emph{et~al.}, ``A real-time and fully distributed approach
  to motion planning for multirobot systems,'' \emph{IEEE Transactions on
  Systems, Man, and Cybernetics: Systems}, 2017.

\bibitem{mora2018cooperative}
J.~Alonso-Mora, P.~Beardsley, and R.~Siegwart, ``Cooperative collision
  avoidance for nonholonomic robots,'' \emph{IEEE Transactions on Robotics},
  vol.~34, no.~2, pp. 404--420, 2018.

\bibitem{mellinger2011minimum}
D.~Mellinger and V.~Kumar, ``Minimum snap trajectory generation and control for
  quadrotors,'' in \emph{Robotics and Automation (ICRA), 2011 IEEE
  International Conference on}.\hskip 1em plus 0.5em minus 0.4em\relax IEEE,
  2011, pp. 2520--2525.

\bibitem{optimalwalambe2016}
R.~Walambe, N.~Agarwal, S.~Kale, and V.~Joshi, ``Optimal trajectory generation
  for car-type mobile robot using spline interpolation∗,''
  \emph{IFAC-PapersOnLine}, vol.~49, no.~1, pp. 601--606, 2016.

\bibitem{Tang2009}
C.~P. Tang, ``Differential flatness-based kinematic and dynamic control of a
  differentially driven wheeled mobile robot,'' in \emph{Robotics and
  Biomimetics (ROBIO), 2009 IEEE International Conference on}.\hskip 1em plus
  0.5em minus 0.4em\relax IEEE, 2009, pp. 2267--2272.

\bibitem{vijayroma}
V.~Arvindh, G.~Aadithya~R, S.~Krishnan, and S.~K, ``Collision-free multi robot
  trajectory optimization in unknown environments using decentralized
  trajectory planning,'' in \emph{Robotics and Manufacturing Automation,2018
  IEEE 4th International Symposium on}.\hskip 1em plus 0.5em minus 0.4em\relax
  IEEE, 2018.

\bibitem{robinson2018efficient}
D.~R. Robinson, R.~T. Mar, K.~Estabridis, and G.~Hewer, ``An efficient
  algorithm for optimal trajectory generation for heterogeneous multi-agent
  systems in non-convex environments,'' \emph{IEEE Robotics and Automation
  Letters}, vol.~3, no.~2, pp. 1215--1222, 2018.

\bibitem{schouweenars2001mixed}
T.~Schouwenaars, B.~De~Moor, E.~Feron, and J.~How, ``Mixed integer programming
  for multi-vehicle path planning,'' in \emph{Control Conference (ECC), 2001
  European}.\hskip 1em plus 0.5em minus 0.4em\relax IEEE, 2001, pp. 2603--2608.

\bibitem{mercado2017mixing}
Y.~Diaz-Mercado and M.~Egerstedt, ``Multirobot mixing via braid groups,''
  \emph{IEEE Transactions on Robotics}, vol.~33, no.~6, pp. 1375--1385, 2017.

\bibitem{Solovey2016finding}
K.~Solovey, O.~Salzman, and D.~Halperin, ``Finding a needle in an exponential
  haystack: Discrete rrt for exploration of implicit roadmaps in multi-robot
  motion planning,'' \emph{The International Journal of Robotics Research},
  vol.~35, no.~5, pp. 501--513, 2016.

\bibitem{fan}
F.~Liu and A.~Narayanan, ``Real time replanning based on a* for collision
  avoidance in multi-robot systems,'' in \emph{Ubiquitous Robots and Ambient
  Intelligence (URAI), 2011 8th International Conference on}.\hskip 1em plus
  0.5em minus 0.4em\relax IEEE, 2011, pp. 473--479.

\bibitem{Hedge2016}
R.~Hegde and D.~Panagou, ``Multi-agent motion planning and coordination in
  polygonal environments using vector fields and model predictive control,'' in
  \emph{Control Conference (ECC), 2016 European}.\hskip 1em plus 0.5em minus
  0.4em\relax IEEE, 2016, pp. 1856--1861.

\bibitem{sutorius2017decentralized}
M.~Sutorius and D.~Panagou, ``Decentralized hybrid control for multi-agent
  motion planning and coordination in polygonal environments,''
  \emph{IFAC-PapersOnLine}, vol.~50, no.~1, pp. 6977--6982, 2017.

\bibitem{van2016online}
R.~Van~Parys and G.~Pipeleers, ``Online distributed motion planning for
  multi-vehicle systems,'' in \emph{Proceedings of the 2016 European Control
  Conference}, 2016, pp. 1580--1585.

\bibitem{2017fast}
D.~Zhou, Z.~Wang, S.~Bandyopadhyay, and M.~Schwager, ``Fast, on-line collision
  avoidance for dynamic vehicles using buffered voronoi cells,'' \emph{IEEE
  Robotics and Automation Letters}, vol.~2, no.~2, pp. 1047--1054, 2017.

\bibitem{Berg2008}
J.~Van~den Berg, M.~Lin, and D.~Manocha, ``Reciprocal velocity obstacles for
  real-time multi-agent navigation,'' in \emph{Robotics and Automation, 2008.
  ICRA 2008. IEEE International Conference on}.\hskip 1em plus 0.5em minus
  0.4em\relax IEEE, 2008, pp. 1928--1935.

\bibitem{Fiorini1998}
P.~Fiorini and Z.~Shiller, ``Motion planning in dynamic environments using
  velocity obstacles,'' \emph{The International Journal of Robotics Research},
  vol.~17, no.~7, pp. 760--772, 1998.

\bibitem{Jur2011}
J.~Van Den~Berg, S.~J. Guy, M.~Lin, and D.~Manocha, ``Reciprocal n-body
  collision avoidance,'' in \emph{Robotics research}.\hskip 1em plus 0.5em
  minus 0.4em\relax Springer, 2011, pp. 3--19.

\bibitem{rufli2013reciprocal}
M.~Rufli, J.~Alonso-Mora, and R.~Siegwart, ``Reciprocal collision avoidance
  with motion continuity constraints,'' \emph{IEEE Transactions on Robotics},
  vol.~29, no.~4, pp. 899--912, 2013.

\bibitem{snape2010smooth}
J.~Snape, J.~Van Den~Berg, S.~J. Guy, and D.~Manocha, ``Smooth and
  collision-free navigation for multiple robots under differential-drive
  constraints,'' in \emph{Intelligent Robots and Systems (IROS), 2010 IEEE/RSJ
  International Conference on}.\hskip 1em plus 0.5em minus 0.4em\relax IEEE,
  2010, pp. 4584--4589.

\bibitem{bareiss2017general}
D.~Bareiss and J.~van~den Berg, ``Generalized reciprocal collision avoidance,''
  \emph{The International Journal of Robotics Research}, vol.~34, no.~12, pp.
  1501--1514, 2015.

\bibitem{cole2018reactive}
K.~Cole and A.~M. Wickenheiser, ``Reactive trajectory generation for multiple
  vehicles in unknown environments with wind disturbances,'' \emph{IEEE
  Transactions on Robotics}, vol.~34, no.~5, pp. 1333--1348, 2018.

\bibitem{hoy2012collision}
M.~Hoy, A.~S. Matveev, and A.~V. Savkin, ``Collision free cooperative
  navigation of multiple wheeled robots in unknown cluttered environments,''
  \emph{Robotics and Autonomous Systems}, vol.~60, no.~10, pp. 1253--1266,
  2012.

\bibitem{ferrera2017decentralized}
E.~Ferrera, J.~Capitan, A.~R. Castano, and P.~J. Marron, ``Decentralized safe
  conflict resolution for multiple robots in dense scenarios,'' \emph{Robotics
  and Autonomous Systems}, vol.~91, pp. 179--193, 2017.

\bibitem{xu2013}
W.-b. Xu, X.-b. Chen, J.~Zhao, and T.-y. Huang, ``A decentralized method using
  artificial moments for multi-robot path-planning,'' \emph{International
  Journal of Advanced Robotic Systems}, vol.~10, no.~1, p.~24, 2013.

\bibitem{Bekris2017Safe}
K.~E. Bekris, D.~K. Grady, M.~Moll, and L.~E. Kavraki, ``Safe distributed
  motion coordination for second-order systems with different planning
  cycles,'' \emph{The International Journal of Robotics Research}, vol.~31,
  no.~2, pp. 129--150, 2012.

\bibitem{safe2018}
H.~Oleynikova, Z.~Taylor, R.~Siegwart, and J.~Nieto, ``Safe local exploration
  for replanning in cluttered unknown environments for microaerial vehicles,''
  \emph{IEEE Robotics and Automation Letters}, vol.~3, no.~3, pp. 1474--1481,
  2018.

\bibitem{2017real}
V.~Usenko, L.~von Stumberg, A.~Pangercic, and D.~Cremers, ``Real-time
  trajectory replanning for mavs using uniform b-splines and a 3d circular
  buffer,'' in \emph{Intelligent Robots and Systems (IROS), 2017 IEEE/RSJ
  International Conference on}.\hskip 1em plus 0.5em minus 0.4em\relax IEEE,
  2017, pp. 215--222.

\bibitem{convexregions}
R.~Deits and R.~Tedrake, ``Computing large convex regions of obstacle-free
  space through semidefinite programming,'' in \emph{Algorithmic foundations of
  robotics XI}.\hskip 1em plus 0.5em minus 0.4em\relax Springer, 2015, pp.
  109--124.

\bibitem{plann}
S.~Liu, M.~Watterson, K.~Mohta, K.~Sun, S.~Bhattacharya, C.~J. Taylor, and
  V.~Kumar, ``Planning dynamically feasible trajectories for quadrotors using
  safe flight corridors in 3-d complex environments,'' \emph{IEEE Robotics and
  Automation Letters}, vol.~2, no.~3, pp. 1688--1695, 2017.

\bibitem{partition}
F.~Altch{\'e} and A.~De~La~Fortelle, ``Partitioning of the free space-time for
  on-road navigation of autonomous ground vehicles,'' in \emph{Decision and
  Control (CDC), 2017 IEEE 56th Annual Conference on}.\hskip 1em plus 0.5em
  minus 0.4em\relax IEEE, 2017, pp. 2126--2133.

\bibitem{gao2018online}
F.~Gao, W.~Wu, Y.~Lin, and S.~Shen, ``Online safe trajectory generation for
  quadrotors using fast marching method and bernstein basis polynomial,'' in
  \emph{2018 IEEE International Conference on Robotics and Automation
  (ICRA)}.\hskip 1em plus 0.5em minus 0.4em\relax IEEE, 2018, pp. 344--351.

\bibitem{diehl2005real}
M.~Diehl, H.~G. Bock, and J.~P. Schl{\"o}der, ``A real-time iteration scheme
  for nonlinear optimization in optimal feedback control,'' \emph{SIAM Journal
  on control and optimization}, vol.~43, no.~5, pp. 1714--1736, 2005.

\bibitem{diehl2002real}
M.~Diehl, H.~G. Bock, J.~P. Schl{\"o}der, R.~Findeisen, Z.~Nagy, and
  F.~Allg{\"o}wer, ``Real-time optimization and nonlinear model predictive
  control of processes governed by differential-algebraic equations,''
  \emph{Journal of Process Control}, vol.~12, no.~4, pp. 577--585, 2002.

\bibitem{vukov2015embedded}
M.~Vukov, ``Embedded model predictive control and moving horizon estimation for
  mechatronics applications,'' 2015.

\bibitem{boyd2004convex}
S.~Boyd and L.~Vandenberghe, \emph{Convex optimization}.\hskip 1em plus 0.5em
  minus 0.4em\relax Cambridge university press, 2004.

\bibitem{bsplinemat}
K.~Qin, ``General matrix representations for b-splines,'' \emph{The Visual
  Computer}, vol.~16, no. 3-4, pp. 177--186, 2000.

\bibitem{pmpc}
C.~Liu, S.~Lee, S.~Varnhagen, and H.~E. Tseng, ``Path planning for autonomous
  vehicles using model predictive control,'' in \emph{Intelligent Vehicles
  Symposium (IV), 2017 IEEE}.\hskip 1em plus 0.5em minus 0.4em\relax IEEE,
  2017, pp. 174--179.

\bibitem{rob}
M.~Kamel, J.~Alonso-Mora, R.~Siegwart, and J.~Nieto, ``Robust collision
  avoidance for multiple micro aerial vehicles using nonlinear model predictive
  control,'' in \emph{Intelligent Robots and Systems (IROS), 2017 IEEE/RSJ
  International Conference on}.\hskip 1em plus 0.5em minus 0.4em\relax IEEE,
  2017, pp. 236--243.

\bibitem{shravanicra}
S.~Krishnan, G.~Aadithya~R, and S.~K, ``Continuous time trajectory optimization
  for decentralized multi robot navigation,'' in \emph{Robotics and Automation
  (ICRA), 2019 IEEE International Conference on}, 2019, \textit{Submitted}.

\bibitem{de1978practical}
C.~De~Boor, \emph{A practical guide to splines}.\hskip 1em plus 0.5em minus
  0.4em\relax Springer-Verlag New York, 1978, vol.~27.

\bibitem{qpoases}
H.~J. Ferreau, C.~Kirches, A.~Potschka, H.~G. Bock, and M.~Diehl, ``qpoases: A
  parametric active-set algorithm for quadratic programming,''
  \emph{Mathematical Programming Computation}, vol.~6, no.~4, pp. 327--363,
  2014.

\bibitem{mpc}
W.~F. Lages and J.~A.~V. Alves, ``Real-time control of a mobile robot using
  linearized model predictive control,'' \emph{IFAC Proceedings Volumes},
  vol.~39, no.~16, pp. 968--973, 2006.

\bibitem{rotors}
F.~Furrer, M.~Burri, M.~Achtelik, and R.~Siegwart, ``Rotors-a modular gazebo
  mav simulator framework,'' in \emph{Robot Operating System (ROS)}.\hskip 1em
  plus 0.5em minus 0.4em\relax Springer, 2016, pp. 595--625.

\bibitem{taecontroller}
T.~Lee, M.~Leoky, and N.~H. McClamroch, ``Geometric tracking control of a
  quadrotor uav on se (3),'' in \emph{Decision and Control (CDC), 2010 49th
  IEEE Conference on}.\hskip 1em plus 0.5em minus 0.4em\relax IEEE, 2010, pp.
  5420--5425.

\bibitem{swept}
W.~Wang and K.~Wang, ``Geometric modeling for swept volume of moving solids,''
  \emph{IEEE Computer graphics and Applications}, vol.~6, no.~12, pp. 8--17,
  1986.

\bibitem{csg}
A.~Requicha and S.~Chan, ``Representation of geometric features, tolerances,
  and attributes in solid modelers based on constructive geometry,'' \emph{IEEE
  Journal on Robotics and Automation}, vol.~2, no.~3, pp. 156--166, 1986.

\end{thebibliography}


\begin{IEEEbiography}[{\includegraphics[width=1in,height=1.25in,clip,keepaspectratio]{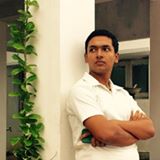}}]{Govind Aadithya R}
graduated in Mechatronics Engineering from SRM Institute of Science and Technology. Furthermore, He was a part of SRM Team Humanoid since September 2015 and led the computer vision domain from May 2017 to June 2018.

His ultimate goal is to develop a Humanoid which is completely aware of its environment through vision based perception and capable of dynamic stabilization. His areas of interest lie in the field of Legged Robotics, mainly in gait planning, stabilization and environment perception through Computer Vision.
\end{IEEEbiography}

\begin{IEEEbiography}[{\includegraphics[width=1in,height=1.25in,clip,keepaspectratio]{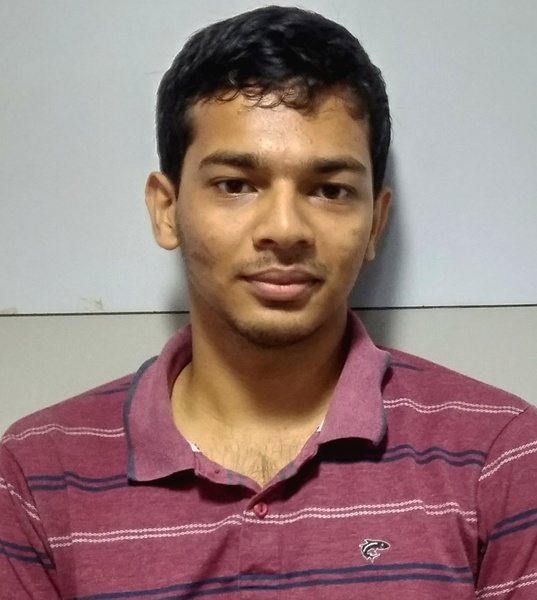}}]{Shravan Krishnan} completed his Bachelors in Mechatronics from SRM Institute of Science and Technology, in India in 2018.

He is currently a research intern at Autonomous Systems Lab, SRM Institute of Science and Technology, Kattankulathur,India. His research interests are decentralized multi-robot navigation in complex and dynamic environments, aerial robots and model predictive control of robotic systems. 
\end{IEEEbiography}


\begin{IEEEbiography}[{\includegraphics[width=1in,height=1.25in,clip,keepaspectratio]{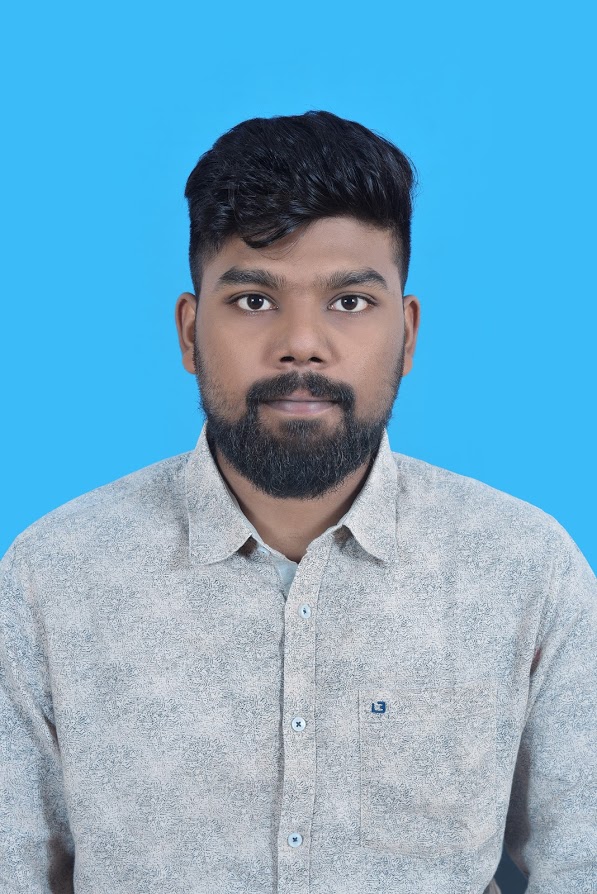}}]{Vijay Arvindh}
is a graduate in B.Tech Mechatronics engineering program at SRM Institute of Science and Technology. 

His areas of interest lie in the overlap of robotics, decentralized control, and distributed algorithms with an affinity towards Autotronics and Auomotive Autonomy. \end{IEEEbiography}

\vfill

\begin{IEEEbiography}
[{\includegraphics[width=1in,height=1.25in,clip,keepaspectratio]{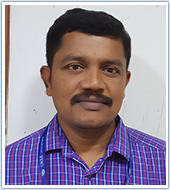}}] {Sivanathan K}received his B.E degree in electronics and communication engineering from K.S.Rangasamy College of Technology, Periyar University, India,  in 2003 and M.E degree in Mechatronics from Madras Institute of Technology campus, Anna University, India, in 2008, and is currently pursuing his Ph.D from SRM Institute of Science and Technology while working as an Assistant Professor at the Department of Mechatronics Engineering and concurrently heads Autonomous Systems lab, a research lab intended for pursuing cutting-edge research in the field of robotics and mechatronics. He is also a faculty advisor for multiple student teams that participate in international robotic competitions. He has twelve years of teaching experience and has published research papers in journals and conferences of international repute. He is proficient in various technical softwares and is a Certified LabVIEW Associate Developer(CLAD) by National Instruments. He is also a Life member of Indian Society for Technical Education (ISTE), Indian Science Congress Association(ISCA) and the IRED.

His research interests include Control of Autonomous Systems - Ground, Aquatic and Aerial robots, Fluid power automation and developing teaching methodologies for effective teaching of Mechatronics engineering.
\end{IEEEbiography}


\enlargethispage{-5in}

\end{document}